\documentclass{article}

\PassOptionsToPackage{numbers, compress}{natbib}
\usepackage[main, final]{neurips_2026}

\usepackage[utf8]{inputenc} 
\usepackage[T1]{fontenc}    
\usepackage[hypertexnames=false]{hyperref}       
\usepackage{url}            
\usepackage{booktabs}       
\usepackage{amsfonts}       
\usepackage{nicefrac}       
\usepackage{microtype}      
\usepackage{wrapfig}
\usepackage{graphicx}
\usepackage{subfigure}
\usepackage{enumitem}
\usepackage{titletoc}

\usepackage{amsmath}
\usepackage{amssymb}
\usepackage{mathtools}
\usepackage{amsthm}
\usepackage{listings}
\usepackage{algorithm}
\usepackage{algorithmic}

\makeatletter
\providecommand*{\toclevel@algorithm}{0}
\makeatother

\usepackage{colortbl}
\usepackage{multirow}
\usepackage{array}
\usepackage{tcolorbox}
\tcbuselibrary{breakable}

\usepackage{tabularx}

\theoremstyle{plain}
\newtheorem{theorem}{Theorem}[section]
\newtheorem{proposition}[theorem]{Proposition}

\theoremstyle{definition}

\newtheorem{assumption}[theorem]{Assumption}
\theoremstyle{remark}

\usepackage[textsize=tiny]{todonotes}

\usepackage{transparent}
\DeclareRobustCommand{\transp}[1]{\transparent{#1}}
\usepackage[dvipsnames]{xcolor}
\definecolor{mediumtealblue}{rgb}{0.0, 0.33, 0.71}
\definecolor{brickred}{rgb}{0.8, 0.25, 0.33}

\title{MaPPO: Maximum a Posteriori Preference Optimization with Prior Knowledge}

\author{
Guangchen Lan\thanks{Equal contribution. Correspondence to: Guangchen Lan \textrm{<lan44@purdue.edu>}} \\
Purdue University \\
\And
Sipeng Zhang$^{*}$ \\
University of California, San Diego
\And
Tianle Wang \\
Purdue University \\
\And
Yuwei Zhang \\
University of California, San Diego
\And
Daoan Zhang \\
University of Rochester \\
\And
Xinpeng Wei \\
Amazon, AGI SF Lab \\
\And
Xiaoman Pan \\
Amazon \\
\And
Hongming Zhang \\
FAIR, Meta \\
\And
Dong-Jun Han \\
Yonsei University \\
\And
Christopher G. Brinton \\
Purdue University \\
}

\begin{document}

\maketitle

\begin{abstract}
As the era of large language models (LLMs) unfolds, Preference Optimization (PO) methods have become a central approach to aligning LLMs with human preferences and improving performance. We propose Maximum a Posteriori Preference Optimization (MaPPO), a methodology for learning from preferences that explicitly incorporates prior reward knowledge into the optimization objective. Building on the paradigm employed by Direct Preference Optimization (DPO) and its variants of treating preference learning as a Maximum Likelihood Estimation (MLE) problem, MaPPO integrates prior reward estimates into a principled Maximum a Posteriori (MaP) objective. This not only generalizes DPO and its variants, but also enhances alignment by mitigating the oversimplified binary classification of responses. Additionally, MaPPO introduces no additional hyperparameters, and supports preference optimization in both offline and online settings. In addition, MaPPO can be used as a plugin for DPO variants, including widely used SimPO, IPO, and CPO, and produce consistent improvements. Extensive empirical evaluations of different model sizes and model series on three standard benchmarks (MT-Bench, AlpacaEval 2.0, and Arena-Hard) demonstrate consistent improvements in alignment performance without sacrificing computational efficiency.
\end{abstract}

\section{Introduction}
\label{Sec:Introduction}

Reinforcement Learning from Human Feedback (RLHF) has shown promise in aligning large language models (LLMs) with human preferences.  Pioneering work framed the problem as RL on a reward model trained from group-wise comparisons, yielding notable improvements in tasks including games~\citep{christiano2017deep} and dialogue~\citep{stiennon2020learning}. Subsequent systems such as InstructGPT~\citep{ouyang2022training} demonstrated that RLHF could scale to billion‑parameter models and substantially boost helpfulness and safety~\citep{lan2025contextual}. Despite its practical successes, RLHF still has a multitude of challenges stemming from noisy feedback, reward-model misalignment, optimization instability, and computation inefficiency with high memory cost, which together hinder its scalability and reliability~\citep{casper2023open, dai2023safe, pan2022effects}.

Direct Preference Optimization (DPO)~\citep{dpo_neurips2023} has addressed some of these issues by reframing the preference learning with a Kullback–Leibler (KL) regularized log-odds objective, effectively turning the task into Maximum Likelihood Estimation (MLE) over pairwise comparisons.
In DPO, the model is trained to assign a higher likelihood to the preferred response than to the rejected, while staying close to a reference policy.
The MLE perspective enhances the efficiency of RLHF, as it eliminates the need for rollouts or value functions.
On the other hand, it introduces a new limitation: by considering only the relative likelihoods within each pair, DPO overlooks the absolute reward magnitude and any external prior knowledge~\citep{amini2024direct, d2025anchored}. 
The training signal in DPO is inherently local, bounded by pairwise comparisons, and lacks global calibration across pre-trained knowledge.

\paragraph{Challenge.}
A fundamental limitation of MLE-based preference optimization lies in its purely relative nature: it focuses on maximizing the \emph{gap} between chosen and rejected responses, yet lacks a mechanism to anchor their absolute probabilities.
As training progresses, the MLE objective tends to suppress the likelihood of the rejected response rather than elevate that of the preferred one. Empirical investigations~\citep{pal2024smaug, rafailov2024r, tajwar2024preference,zhang2024negative} consistently show a simultaneous reduction in the absolute probabilities assigned to both preferred and rejected answers, resulting in abnormal output distributions.
This undesirable dynamic, empirically known as the \textit{squeezing effect}~\citep{ren2024learning}, undermines policy calibration and injects instability into generation. 
The issue is especially severe in near-tie cases (see Figure~\ref{fig:compare_MLE_MAP}), where models approach human-level performance~\citep{liu2024reward, guo2024todo}. Both responses are reasonable yet MLE still enforces an artificial separation, draining probability mass from the high-quality region of the output space. 
Motivated by this, the key question that this paper aims to address is:



{\em How can we improve language model alignment through a more principled training signal, instead of an oversimplified MLE pipeline?}

\paragraph{Overview of approach.} In this paper, we answer the above question by introducing Maximum-a-Posteriori Preference Optimization (MaPPO), a simple yet principled extension of DPO that injects data‑driven prior knowledge into preference training. 
MaPPO augments the standard maximum-likelihood objective with a lightweight MaP regularizer -- an additional log‑prior scaled by a calibrated reward gap -- which proportionally adjusts each update to the confidence differential between the preferred and rejected answers. 
Instead of the oversimplified binary classification in MLE, this mechanism curbs the excessive penalization of near‑tie pairs while preserving DPO’s closed form and computational efficiency.
Moreover, the proposed MaPPO is suitable for both offline and online settings.
Furthermore, MaPPO is designed as a drop-in regularization module and seamlessly integrates with a broad spectrum of recent DPO variants.

Extensive experiments demonstrate that MaPPO delivers consistently stronger performance across three public alignment benchmarks: AlpacaEval 2.0, Arena-Hard, and MT-Bench. 
We evaluate MaPPO on multiple model families, including Llama-3, Qwen2.5, and Mistral, under multiple model sizes.
Compared with DPO on Mistral-7B-Instruct, MaPPO improves the win rate on AlpacaEval 2.0 from $18.24$ to $30.56$, an absolute gain of $12.32$ points, and improves Arena-Hard from $14.2$ to $18.4$, an absolute gain of $4.2$ points.
These results validate that a lightweight prior is sufficient to produce stronger and better-calibrated policies.
Furthermore, we evaluate MaPPO integrated with Iterative-DPO~\citep{dong2024rlhf}, SimPO~\citep{meng2024simpo}, IPO~\citep{pmlr-v238-gheshlaghi-azar24a}, and CPO~\citep{xu2024contrastive} in Appendix \ref{app:Illustration} and \ref{app:adapt}. 
In all cases, we observe consistent gains up to $31.3\%$ on Arena-Hard in alignment scores without requiring additional computation or changes to the optimization pipeline. 
This suggests that MaPPO serves as a robust and general enhancement strategy for advanced preference training pipelines.

\paragraph{Contributions.} 
In summary, the main contributions of this work are as follows:
\begin{enumerate}[leftmargin=12pt]
\item We propose MaPPO, a principled generalization of Direct Preference Optimization, which incorporates data-driven prior reward estimates into a Maximum-a-Posteriori (MaP) objective.
\item We demonstrate that MaPPO naturally supports both offline (e.g., DPO) and online (e.g., Iterative-DPO) preference optimization.
\item We show that MaPPO is \textbf{compatible} with and enhances existing DPO variants, including SimPO, IPO, and CPO. For all variants, no additional hyperparameters are needed.
\item Experiments across multiple model series and model sizes confirm consistent improvements in alignment performance on standard benchmarks, including MT-Bench, AlpacaEval 2.0, and Arena-Hard.
\end{enumerate}


\section{Preliminaries and Problem Setup}
\label{Sec:Problem_Setup}

\subsection{RL Tuning}

First, we introduce the general framework of Reinforcement Learning (RL). Consider the Markov decision process (MDP) as a tuple $( \mathcal{S}, \mathcal{A}, \mathcal{P}, \mathcal{R})$, where $\mathcal{S}$ is the state space, $\mathcal{A}$ is a finite action space, $\mathcal{P}:\mathcal{S}\times\mathcal{A}\times\mathcal{S}\rightarrow\mathbb{R}$ is a Markov kernel that determines transition probabilities, and $\mathcal{R}:\mathcal{S}\times\mathcal{A}\rightarrow\mathbb{R}$ is a reward function. 
At each time step $t$, the agent executes an action $\mathbf{y}_t \in\mathcal{A}$ from the current state $\mathbf{s}_t \in\mathcal{S}$, following a stochastic policy $\pi$, \textit{i.e.}, $\mathbf{y}_t \sim \pi(\cdot|\mathbf{s}_t)$. 
The corresponding reward is defined as $r_t$.

Following the conventional setting in LLMs, the policy $\pi_{\theta}$ represents the LLM with model parameters $\theta$. The action space $\mathcal{A}$ is set as the vocabulary. At step $t$, $\mathbf{s}_t=(\mathbf{x}, \mathbf{y}_{<t})$ is a cascade of the query $\mathbf{x}$ and the tokens $\mathbf{y}_{<t} = (\mathbf{y}_{1}, \cdots, \mathbf{y}_{t-1})$ that have been predicted, and $\mathbf{y}_t$ is the next token to be predicted. The transition kernel $\mathcal{P}$ is deterministic as $\mathbf{s}_{t+1} = (\mathbf{s}_t, \mathbf{y}_t)$. The complete answer $\mathbf{y}=(\mathbf{y}_{1}, \cdots,\mathbf{y}_{T})$ with length $|\mathbf{y}|=T$. The step reward $r_t = r(\mathbf{x}, \mathbf{y}_{\le t})$ can be obtained from a trained reward model.

After formalizing the LLM tuning as an RL problem, the goal of RL tuning~\citep{ouyang2022training} is to maximize the expectation of the cumulative reward $r \coloneqq r(\mathbf{x}, \mathbf{y})$  with a Kullback–Leibler (KL) constraint as follows
\begin{equation}
\label{eq:rl_target}
\begin{aligned}
    \max_{\pi_{\theta}} \mathop{\mathbb{E}}_{\mathbf{x}\sim\mathcal{D},\atop \mathbf{y}\sim\pi_{\theta}(\cdot|\mathbf{x})} \left[ r(\mathbf{x}, \mathbf{y}) \right] - \beta \mathbb{D}_{\rm KL} \left[ \pi_{\theta}(\cdot|\mathbf{x})||\pi_{\rm ref}(\cdot|\mathbf{x}) \right],
\end{aligned}
\end{equation}
where $\mathbb{D}_{\rm KL}(\cdot || \cdot)$ denotes the KL divergence, and $\beta$ is a constant weight. $\pi_{\rm ref}$ is a reference model, which is usually the initial policy model before tuning. This optimization problem can be solved by any RL algorithms, e.g., PPO \citep{schulman2017proximal} and GRPO \citep{shao2024deepseekmath}.

\subsection{Direct Preference Optimization}

In Direct Preference Optimization (DPO) \citep{dpo_neurips2023}, a closed-form expression of \eqref{eq:rl_target} is used, and a connection between policy $\pi$ and reward function $r$ is built as
\begin{equation}
\label{eq:r_pi_closed}
\begin{aligned}
    \pi( \mathbf{y} | \mathbf{x} ) ~=~& 
    \frac{1}{Z(\mathbf{x})} \pi_{\rm ref}( \mathbf{y} | \mathbf{x} ) \exp \big( \frac{1}{\beta} r(\mathbf{y}, \mathbf{x}) \big),
\end{aligned}
\end{equation}
where $Z(\mathbf{x}) = \sum_{\mathbf{y}} \pi_{\mathrm{ref}}(\mathbf{y} \mid \mathbf{x}) \exp\left(\frac{1}{\beta} r(\mathbf{x}, \mathbf{y})\right)$ is a partition function to normalize the probability and irrelevant to $\theta$.

With a prompt $\mathbf{x}$, we sample two responses from the current policy model $\mathbf{y}_{1}, \mathbf{y}_{2} \sim\pi_{\theta}(\cdot|\mathbf{x})$. 
A human expert then demonstrates the preference and ranks the responses as $\mathbf{y}_{w}$ (win) and $\mathbf{y}_{l}$ (lose). 
After plugging in \eqref{eq:r_pi_closed} into the reward model training (MLE) loss function, the target of RL tuning becomes to minimize the loss function shown below
\begin{equation}
\label{eq:dpo_loss}
\begin{aligned}
    \mathcal{L}(\theta) ~=~&
    \mathop{\mathbb{E}}_{( \mathbf{y}_{w}, \mathbf{y}_{l}, \mathbf{x})\sim\mathcal{D}} \Big[ -\log \sigma \Big( \beta\log \frac{\pi_{\theta}(\mathbf{y}_{w}|\mathbf{x})}{\pi_{\rm ref}(\mathbf{y}_{w}|\mathbf{x})} - \beta\log \frac{\pi_{\theta}(\mathbf{y}_{l}|\mathbf{x})}{\pi_{\rm ref}(\mathbf{y}_{l}|\mathbf{x})} \Big)\Big],
\end{aligned}
\end{equation}
where $\sigma (\cdot)$ is the logistic function. The training process can be done using mini-batch gradient descent and the variants, e.g., AdamW \citep{loshchilov2018decoupled}. Notably, the derivation on reward model training is based on MLE, which oversimplifies the process as a binary classification problem. As a result, minimizing \eqref{eq:dpo_loss} is nothing but increasing the gap between the output probability of $\mathbf{y}_{w}$ and $\mathbf{y}_{l}$.

\subsection{Current issues with MLE}

The learning process is to estimate the parameters $\theta$ via maximum likelihood, which is to maximize the gap of the output probability between the winning response $\pi_{\theta}(\mathbf{y}_{w}|\mathbf{x})$ and the losing response $\pi_{\theta}(\mathbf{y}_{l}|\mathbf{x})$.

\begin{figure}[!htbp]
    \centering  \includegraphics[width=0.75\linewidth]{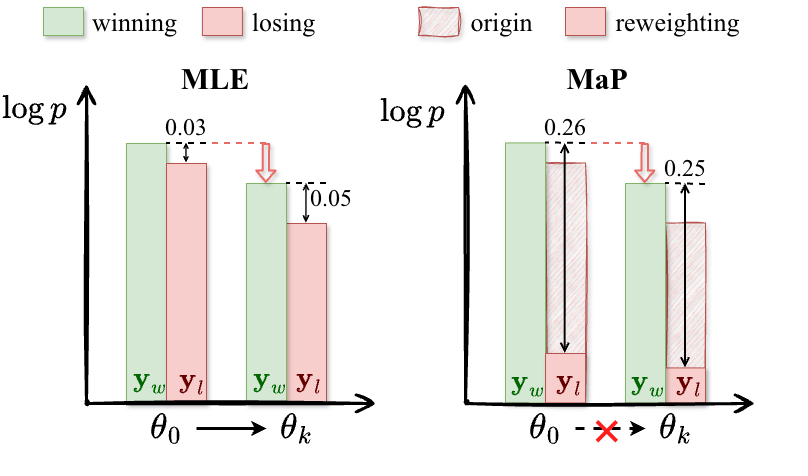}
    \caption{Under the standard MLE-based DPO (left), empirical studies~\citep{pal2024smaug, rafailov2024r, tajwar2024preference, ren2024learning} demonstrated that training tends to simultaneously downscale (with different magnitudes) both the chosen and rejected responses to increase their gap. Our MaP-based method (right) mitigates this harmful tendency by re-weighting the rejected response based on prior knowledge. Here, the x-axis denotes the initial model $\theta_0$ and a potentially harmful model $\theta_k$ that may arise during training, while the y-axis shows the log-likelihood of a fixed preference pair under different policies.}
\label{fig:compare_MLE_MAP}
\end{figure}

Despite its computational efficiency and simplicity, the MLE formulation introduces a critical limitation: It focuses solely on relative probabilities within each preference pair, while ignoring the absolute magnitude of confidence in the responses. Shown as Figure~\ref{fig:compare_MLE_MAP}, training encourages simultaneous downscaling of both $\pi_{\theta}(\mathbf{y}_{w}|\mathbf{x})$ and $\pi_{\theta}(\mathbf{y}_{l}|\mathbf{x})$, to enlarge the preference gap. This undesired tendency can lead to over-penalization of both responses, especially in near-tie cases, thereby reducing output confidence and harming policy calibration~\citep{ren2024learning}.

Such issues become particularly problematic as models approach human-level performance, where both the winning and losing responses may be of high quality. For example, consider a preference pair where both $\mathbf{y}_{w}$ and $\mathbf{y}_{l}$ are grammatically correct, contextually relevant, and factually accurate, with the only preference driven by stylistic nuances. In this case, the MLE objective may still enforce a large probability gap by downscaling both log-likelihoods. This inadvertently shifts the probability mass away from the high-quality response space, introducing unnecessary uncertainty into the policy, and can lead to erratic or unstable generation in downstream usage.

\begin{figure}[!ht]
\begin{tcolorbox}[title=Prompt~$\mathbf{x}$, fonttitle=\bfseries, boxrule=0.5mm]
A printer prints 17 pages per minute. How many minutes will it take to print 200 pages? Express your answer to the nearest whole number.
\end{tcolorbox}
\begin{tcolorbox}[colback=green!5, colframe=black!60!green!45, title={\textbf{Response}~$\mathbf{y}_{w}$, $r=0.95$}, fonttitle=\bfseries, boxrule=0.5mm]
To find out how many minutes it will take to print 200 pages, we need to divide the total number of pages by the number of pages the printer can print per minute.
So we divide 200 by 17: $\frac{200}{17} \approx 11.76$.
But since we want the nearest whole number, we need to round it up. That means it will take $\boxed{12}$ minutes to print 200 pages. The answer is: $\boxed{12}$
\end{tcolorbox}
\begin{tcolorbox}[colback=red!5, colframe=black!60!red!45, title={\textbf{Response}~$\mathbf{y}_{l}$, $r=0.91$}, fonttitle=\bfseries, boxrule=0.5mm]
To find out how many minutes it will take to print 200 pages at a rate of 17 pages per minute, divide: $\frac{200}{17} \approx 11.76$.
Rounding 11.76 to the nearest whole number gives: $12$.
So it will take $\boxed{12}$ minutes to print 200 pages. The answer is: $\boxed{12}$
\end{tcolorbox}
\caption{An example of $(\mathbf{x}, \mathbf{y}_{w}, \mathbf{y}_{l})$ pair. Both responses $\mathbf{y}_{w}$ and $\mathbf{y}_{l}$ have good quality as they achieve high rewards, where $r(\mathbf{x}, \mathbf{y}_{w}) = 0.95$, $r(\mathbf{x}, \mathbf{y}_{l}) = 0.91$, and $r \in [0,1]$.}
\label{fig:prompt_example}
\end{figure}

\begin{table}[!ht]
\centering
\caption{Comparison of log-probabilities for responses $\mathbf{y}_w$ and $\mathbf{y}_l$ in PO methods.}
\label{table:logp_gap_comparison}
\begin{tabular}{lllc}
\toprule
{Model} & $\log p(\mathbf{y}_w)$ & $\log p(\mathbf{y}_l)$ & $\log p(\mathbf{y}_w)$-$\log p(\mathbf{y}_l)$ \\
\midrule
IT         & $-14.3$    & $-43.4$    & $+29.1$ \\
DPO        & \textcolor{red}{$-121.5$}   & $-443.2$   & $+321.7$ \\
MaPPO      & \textcolor{black!50!green!80}{$-10.4$}    & $-321.1$   & $+310.7$ \\
\bottomrule
\end{tabular}
\end{table}

In Figure \ref{fig:prompt_example}, we show an example of the prompt and response pair $(\mathbf{x}, \mathbf{y}_{w}, \mathbf{y}_{l})$.  
Given the prompt $\mathbf{x}$, both $\mathbf{y}_{w}$ and $\mathbf{y}_{l}$ have correct answers, while in slightly different expressions. 
Both $\mathbf{y}_{w}$ and $\mathbf{y}_{l}$ have high rewards, which reflect the high qualities.
However, in MLE and its derived DPO, the learning objective is nothing but to increase the gap between $\mathbf{y}_{w}$ and $\mathbf{y}_{l}$, regardless of the fact that both of them have high qualities with correct answers, and their qualities match each other. 
We also list an example with long responses in Appendix \ref{app:Supplementary_Examples}.
As shown in Table \ref{table:logp_gap_comparison}, after DPO training, the log output probability of $\mathbf{y}_{w}$ decreases from $-14.3$ to $-121.5$, and $\mathbf{y}_{l}$ decreases from $-43.4$ to $-443.2$. Although the gap indeed increases from $29.1$ to $321.7$, it violates the principal goal: Increase the output probability of the high-quality responses.

These issues highlight the need for a more principled formulation that preserves relative preferences while incorporating global calibration and prior reward knowledge. In the next section, we introduce our Maximum-a-Posteriori (MaP) framework that addresses these shortcomings in a unified and efficient manner.

\section{MaPPO Design}
\label{Sec:mappo}

\subsection{MaPPO Loss}
\label{Sec:mappo_loss}

In this subsection, we start the derivation step by step from the first principle. 

With a prompt $\mathbf{x}$ and responses $(\mathbf{y}_{1}, \mathbf{y}_{2})$, an oracle gives its preference on the responses as $(\mathbf{y}_{w}, \mathbf{y}_{l})$. The Bradley-Terry (BT) model \citep{bradley1952rank} builds the connection between the rewards and the preference probability as follows:
\begin{equation}
\label{eq:bt_model}
\begin{aligned}
    p(\mathbf{y}_{w} \succ \mathbf{y}_{l} | \mathbf{x}) ~=~ \frac{ \exp( r(\mathbf{y}_{w}, \mathbf{x}) ) }{\exp( r(\mathbf{y}_{w}, \mathbf{x}) ) + \exp( r(\mathbf{y}_{l}, \mathbf{x}) )} 
    ~=~ \frac{1}{1 + \exp( r(\mathbf{y}_{l}, \mathbf{x}) - r(\mathbf{y}_{w}, \mathbf{x}) )}.
\end{aligned}
\end{equation}
The preference dataset has $N$ samples denoted as $\mathcal{D} = \{\mathbf{y}_{w}^{i}, \mathbf{y}_{l}^{i}, \mathbf{x}^{i}\}_{i=1}^{N}$. We can parametrize a reward model with model parameters $\phi$ as $r_{\phi}(\mathbf{y}, \mathbf{x})$. Given $\mathbf{x}$, assume we have prior knowledge of rewards as $r_{w}$ and $r_{l} \in [-1,0]$. This can be obtained from an oracle, \textit{e.g.}, a pre-trained reward model. To incorporate the prior knowledge of rewards, we need to use the gap $\Delta_{r} = r_{w} - r_{l} \in [0,1]$ as suggested in \eqref{eq:bt_model}. To keep the softmax form in the BT model, we can construct the prior probability as follows
\begin{equation}
\label{eq:prior_prob}
\begin{aligned}
p(r_{\phi}) ~=~ \frac{ \exp( r_{\phi}(\mathbf{y}_{w}, \mathbf{x}) ) + \exp( r_{\phi}(\mathbf{y}_{l}, \mathbf{x}) )}{ \exp( r_{\phi}(\mathbf{y}_{w}, \mathbf{x}) ) + \exp( \Delta_{r} r_{\phi}(\mathbf{y}_{l}, \mathbf{x}) )}.
\end{aligned}
\end{equation}

We use the reward gap $\Delta_{r}$ on the softmax probability to make the probability always greater than $0$ and smaller than $1$. 
Notably, this form is not unique, and other forms are also acceptable if they satisfy the properties of the probability function. We further discuss the prior function in Appendix \ref{app:Prior_Function}.

The MaP loss is the combination of the MLE loss and the prior knowledge loss as follows
\begin{equation}
\label{eq:map_mle_loss}
\begin{aligned}
    \mathcal{L}_{\rm MaP}(r_{\phi}) ~=~& \mathcal{L}_{\rm MLE}(r_{\phi}) + \mathcal{L}_{\rm p}(r_{\phi}) \\
    ~=~& \mathop{\mathbb{E}}_{( \mathbf{y}_{w}, \mathbf{y}_{l}, \mathbf{x})\sim\mathcal{D}} \Big[ -\log \sigma \Big( r_{\phi}(\mathbf{y}_{w}, \mathbf{x}) - r_{\phi}(\mathbf{y}_{l}, \mathbf{x})\Big) - \log p(r_{\phi})\Big] \\
    ~=~& \mathop{\mathbb{E}}_{( \mathbf{y}_{w}, \mathbf{y}_{l}, \mathbf{x})\sim\mathcal{D}} \Big[ -\log \sigma \Big(  r_{\phi}(\mathbf{y}_{w}, \mathbf{x}) - \Delta_{r} r_{\phi}(\mathbf{y}_{l}, \mathbf{x})\Big)\Big].
\end{aligned}
\end{equation}


As proved in previous works \citep{dpo_neurips2023, go2023aligning, korbak2022on}, we have a closed-form connection between the policy $\pi$ and an implicit reward function $r(\mathbf{y}, \mathbf{x})$ as
\begin{equation}
\label{eq:r_pi}
\begin{aligned}
    r(\mathbf{y}, \mathbf{x}) ~=~& \beta \log \big(\frac{\pi( \mathbf{y} | \mathbf{x} )}{\pi_{\rm ref}( \mathbf{y} | \mathbf{x} )} \big).
\end{aligned}
\end{equation}

With a parametrized policy $\pi_{\theta}$, we can plug this result into the loss function \eqref{eq:map_mle_loss}, and get the MaPPO loss
\begin{equation}
\label{eq:mappo_loss}
\begin{aligned}
    \mathcal{L}_{\rm MaP}(\theta) ~=~ 
    \mathop{\mathbb{E}}_{( \mathbf{y}_{w}, \mathbf{y}_{l}, \mathbf{x})\sim\mathcal{D}} \Big[ -\log \sigma \Big( \beta\log \frac{\pi_{\theta}(\mathbf{y}_{w}|\mathbf{x})}{\pi_{\rm ref}(\mathbf{y}_{w}|\mathbf{x})} - \Delta_{r} \beta\log \frac{\pi_{\theta}(\mathbf{y}_{l}|\mathbf{x})}{\pi_{\rm ref}(\mathbf{y}_{l}|\mathbf{x})} \Big)\Big].
\end{aligned}
\end{equation}

With the MaP estimation, we achieve a clean result compared to the MLE estimation in DPO with a calibration term $\Delta_{r} \in [0,1]$ from the prior knowledge.

\paragraph{Remark.}
In our MaPPO method, \textbf{no} additional hyperparameter is introduced compared to the original DPO method. Thus, MaPPO offers a clean and easily pluggable solution, and no extra hyperparameter tuning is needed.

\subsection{Analysis of MaPPO}

In this subsection, we analyze the connection with MaPPO and DPO.

\paragraph{Connection with DPO.} In \eqref{eq:mappo_loss}, when $r_{w}=0$ and $r_{l}=-1$, we have $\Delta_{r}=1$. The loss function becomes
\begin{equation}
\label{eq:mappo_loss_1}
\begin{aligned}
    \mathcal{L}(\theta) ~=~ 
    \mathop{\mathbb{E}}_{( \mathbf{y}_{w}, \mathbf{y}_{l}, \mathbf{x})\sim\mathcal{D}} \Big[ -\log \sigma \Big( \beta\log \frac{\pi_{\theta}(\mathbf{y}_{w}|\mathbf{x})}{\pi_{\rm ref}(\mathbf{y}_{w}|\mathbf{x})} - \beta\log \frac{\pi_{\theta}(\mathbf{y}_{l}|\mathbf{x})}{\pi_{\rm ref}(\mathbf{y}_{l}|\mathbf{x})} \Big)\Big],
\end{aligned}
\end{equation}
which degenerates to the DPO loss function in \eqref{eq:dpo_loss}. 

{Overall, DPO loss function can be taken as \emph{special cases} of MaPPO. In this sense, MaPPO can be taken as a dynamic weighted mechanism, where the weight depends on the relative quality (rewards) of the winning response $\mathbf{y}_{w}$ and the losing response $\mathbf{y}_{l}$.}

\paragraph{Gradient Dynamics Analysis.} To analyze the update of MaPPO, the gradient of \eqref{eq:mappo_loss} is
\begin{equation}
\label{eq:mappo_grad}
\begin{aligned}
    \nabla \mathcal{L}_{\rm MaP}(\theta) ~=~
    \mathop{\mathbb{E}}_{( \mathbf{y}_{w}, \mathbf{y}_{l}, \mathbf{x})\sim\mathcal{D}} \Big[ -\beta (1-\sigma(u)) \Big( \nabla \log {\pi_{\theta}(\mathbf{y}_{w}|\mathbf{x})} - \Delta_{r} \nabla \log {\pi_{\theta}(\mathbf{y}_{l}|\mathbf{x})} \Big)\Big],
\end{aligned}
\end{equation}
where $u = \beta \big( \log \frac{\pi_{\theta}(\mathbf{y}_{w}|\mathbf{x})}{\pi_{\rm ref}(\mathbf{y}_{w}|\mathbf{x})} - \Delta_{r} \log \frac{\pi_{\theta}(\mathbf{y}_{l}|\mathbf{x})}{\pi_{\rm ref}(\mathbf{y}_{l}|\mathbf{x})} \big)$ serves as a confidence measure of preference separation. 
$1-\sigma(u)$ down-weights the gradient when the model is already confident in distinguishing $\mathbf{y}_{w}$ and $\mathbf{y}_{l}$. 
$\Delta_{r}$ directly scales the contribution of the losing sample $\mathbf{y}_{l}$, modulating the penalization. 
The gradient norm of MaPPO is upper-bounded compared to DPO, leading to less aggressive updates and more stable policy calibration. 
We provide a more detailed theoretical analysis, including the stationary convergence analysis in Appendix \ref{app:theo_converge} and Lipschitz stability analysis in Appendix \ref{app:Lipschitz}.

\subsection{Online MaPPO}
\label{sec:online_mappo}

Beyond the offline setting, our MaPPO method can be directly used in the online or iterative settings. As shown in Algorithm \ref{algo:depo}, we describe the online version of MaPPO. In online MaPPO, one key difference is that the responses $\{\mathbf{y}\}$ are generated online from the current policy $\pi_{\theta}$ instead of the initial policy $\pi_{\theta_{0}}$ in the offline setting.

\begin{algorithm}[!ht]
\caption{Online MaPPO}
\label{algo:depo}
\begin{algorithmic}[1] 
\REQUIRE {Prompt data set $\mathcal{D}$; Number of iterations $K$; Initial policy model ${\theta_{0}}$.}
\STATE {\textbf{for} $k = 0, \cdots, K-1$ \textbf{do}}
\STATE {~~~~~~~~Sample a prompt $\mathbf{x} \sim\mathcal{D}$.}
\STATE {~~~~~~~~Sample responses from the current policy $\mathbf{y}_{w}, \mathbf{y}_{l} \sim \pi_{\theta_{k}}(\cdot | \mathbf{x})$.}
\STATE {~~~~~~~~Get corresponding rewards $r_{w} \leftarrow r(\mathbf{y}_{w}, \mathbf{x})$ and $r_{l} \leftarrow r(\mathbf{y}_{l}, \mathbf{x})$.}
\STATE {~~~~~~~~$\Delta_{r} \leftarrow r(\mathbf{y}_{w}, \mathbf{x}) - r(\mathbf{y}_{l}, \mathbf{x})$}
\STATE {~~~~~~~~Compute $\mathcal{L}(\theta_{k})$ according to \eqref{eq:mappo_loss}.}
\STATE {~~~~~~~~$\theta_{k+1} \leftarrow \theta_{k} - \eta \nabla \mathcal{L}(\theta_{k})$~~~~\# Or other optimizer, e.g., AdamW.}
\STATE {\textbf{end for}}
\ENSURE
{${\theta_{K}}$}
\end{algorithmic}
\end{algorithm}

In practice, considering training efficiency, online PO can be implemented in an iterative way, known as I-DPO \citep{dong2024rlhf}. We illustrate the pipeline of iterative DPO in Appendix \ref{app:Illustration}.
With a prompt set $\mathcal{D}$, we can equally divide $\mathcal{D}$ into $K$ subsets as $\mathcal{D}_{1} \cdots \mathcal{D}_{K}$. 
In the $k$-th iteration, we first freeze the current policy model $\pi_{\theta}$, and then get responses $(\mathbf{y}_{1}, \mathbf{y}_{2})$ from the policy according to the prompt set $\mathcal{D}_{k}$. 
We then use a reward model to get the responses' corresponding rewards and collect $(\mathbf{y}_{w}, \mathbf{y}_{l})$ pairs, which reflect the preference. 
After response collection on $\mathcal{D}_{k}$, we conduct the MaPPO training process using \eqref{eq:mappo_loss} on the subset $\mathcal{D}_{k}$. 
After training on the prompt subset, we repeat the process in the next iteration $k+1$ until we finish all $K$ training iterations in MaPPO.

\paragraph{Remark.}
A reward model (or rule-based verifier) is necessary for all online methods, including online DPO (iterative DPO) \citep{dong2024rlhf} and reject sampling \citep{dong2023raft}.

\section{Experiments}
\label{Sec:Experiments}


\subsection{Setup}
\label{sec:Experiments_setup}

\paragraph{Pipeline Settings.}
We follow the RLHF framework in \cite{dong2024rlhf}. 
Instead of costly human annotations, we employ off-the-shelf reward models to generate the preferences. We use the public pre-trained BT reward model\footnote{https://huggingface.co/sfairXC/FsfairX-LLaMA3-RM-v0.1} as the prior knowledge. For the response selection, we follow the rejection sampling strategy suggested by \cite{liustatistical, gulcehre2023reinforced}. 
For each prompt, we generate $n = 8$ responses and use the best-of-8 as $\mathbf{y}_{w}$ and the worst-of-8 as $\mathbf{y}_{l}$. We provide hyperparameter details and computing resources in Appendix \ref{sec:app_exp_setting}.

\paragraph{Dataset.}
We use the prompt set from UltraFeedback~\citep{cui2024ultrafeedback,dong2024rlhf}, which is publicly available\footnote{https://huggingface.co/datasets/RLHFlow/UltraFeedback-preference-standard}. In the offline setting, we generate responses from the initial model with the whole prompt set. In the online (iterative) setting, we separate the prompt set into three subsets of the same size. 
The learning process lasts for $K = 3$ iterations. 
In each iteration, we sample responses from our current policy with one prompt subset, and use preference signals on these responses to improve our policy.

\paragraph{Models.}
To show the scalability of our methods, we choose models in two dimensions: (1) \textit{Model sizes}: Qwen2.5-\{1.5B, 3B, 7B, 14B, 32B\}-Instruct. (2) \textit{Model series}: Qwen2.5-7B-Instruct~\citep{yang2024qwen2}, Mistral-7B-Instruct-v0.3~\citep{jiang2024mistral}, and Llama-3-8B-Instruct~\citep{llama3} in our experiments.

\paragraph{Evaluation.}
We evaluate the model performance on three widely used benchmarks: MT-Bench \citep{zheng2023judging},
AlpacaEval 2.0 \citep{li2023alpacaeval}, and Arena-Hard v0.1 \citep{li2024live}. MT-Bench contains $80$ questions from $8$ categories, with answers rated by GPT-4 on a scale of $1-10$. Arena-Hard v0.1 contains $500$ technical problem-solving questions, and the answers are compared to reference responses from the baseline model GPT-4-0314. We report the win rate (WR) in percentage as judged by GPT-4 Turbo (Preview-1106). AlpacaEval 2.0 includes $805$ questions from five datasets, with the judge model GPT-4 Turbo (Preview-1106) comparing the answers to reference responses from itself. We report the length-controlled (LC) WR as suggested in \cite{dubois2024lengthcontrolled}.

\subsection{Main Results}
\label{sec:Experiments:Main_Results}

Our main results on three standard benchmarks, introduced in Section \ref{sec:Experiments_setup}, are shown in Table \ref{table:Evaluation}. 
For the alignment methods, we show the evaluation results of Instruction Tuning (IT), the original offline setting~\citep{dpo_neurips2023} (DPO), and the online setting~\citep{dong2024rlhf} (I-DPO) as described in Section \ref{sec:online_mappo}. 
For DPO and I-DPO, we show their improvements that incorporate the MaPPO design (+MaPPO). 

For all Llama-3-8B-Instruct, Mistral-7B-Instruct, and Qwen2.5-7B-Instruct models, the performances are significantly improved with MaPPO on AlpacaEval 2.0 and Arena-Hard v0.1 in both the offline setting (DPO) and the online setting (I-DPO). 
It reflects the effectiveness of MaPPO on different model series in both online and offline settings. 
On the MT-Bench, the performances are slightly improved on Qwen2.5-7B-Instruct and Llama-3-8B-Instruct, because the base models have already achieved very good results on this benchmark, which has limitations to reflect the effective improvement. 
The improvement on MT-Bench becomes much more significant on models with mediocre base or DPO performances, \textit{e.g.}, Mistral-7B-Instruct and Qwen2.5-1.5B-Instruct. 
In one model series, the larger models achieve higher overall scores in both base performances and after online \& offline alignment tuning, suggesting that scaling up model size enhances alignment capability as expected. 
With MaPPO, the improvement is consistent in scale with different model sizes in both online and offline settings, and the alignment can make smaller models outperform larger base models.

For AlpacaEval 2.0, the standard errors are consistently small, ranging from $1.02$ to $1.67$ points across model sizes.
$1.02$ occurs on the small model Qwen2.5-1.5B-Instruct, and $1.67$ occurs on the large 32B model.
For Arena-Hard, the confidence intervals are narrow, typically about $(-1.5, +1.5)$ points from 1.5B to 32B models. 
For MT-Bench, our reported number is the average of $3$ judge scores.
Importantly, our main improvements on AlpacaEval 2.0 and Arena-Hard are substantially larger than these uncertainty ranges, for example, on Qwen2.5-7B, MaPPO improves DPO by $+6.23$ on AlpacaEval 2.0 and $+13.7$ on Arena-Hard, and improves I-DPO by $+5.30$ and $+14.7$, respectively.

\begin{wraptable}{r}{0.55\columnwidth}
\centering

\scriptsize
\caption{
Evaluation results of DPO variants with the MaPPO plugin.
$\uparrow$ indicates higher is better.
}
\label{table:adaptation_main}


\renewcommand{\arraystretch}{1}
\setlength{\tabcolsep}{3pt}

\resizebox{0.55\columnwidth}{!}{
\begin{tabular}{l|ccc}
\toprule[1.0pt]

 Method
& AlpacaEval 2.0 $\uparrow$
& Arena-Hard $\uparrow$
&  MT-Bench $\uparrow$ \\

\midrule[1.0pt]

DPO
& 32.01 & 45.5 & 8.56 \\

+MaPPO
& \textbf{38.24} {\scriptsize\textcolor{ForestGreen}{(+6.23)}}
& \textbf{59.2} {\scriptsize\textcolor{ForestGreen}{(+13.7)}}
& \textbf{8.79} {\scriptsize\textcolor{ForestGreen}{(+0.23)}} \\

\cline{1-4}

I-DPO
& 33.80 & 46.9 & \textbf{8.55} \\

+MaPPO
& \textbf{39.10} {\scriptsize\textcolor{ForestGreen}{(+5.30)}}
& \textbf{61.6} {\scriptsize\textcolor{ForestGreen}{(+14.7)}}
& 8.54 {\scriptsize\textcolor{Apricot}{(-0.01)}} \\

\cline{1-4}

SimPO
& 25.15 & 64.2 & \textbf{9.02} \\

+MaPPO
& \textbf{32.75} {\scriptsize\textcolor{ForestGreen}{(+7.60)}}
& \textbf{69.5} {\scriptsize\textcolor{ForestGreen}{(+5.3)}}
& 8.94 {\scriptsize\textcolor{Apricot}{(-0.08)}} \\

\cline{1-4}

IPO
& 27.76 & 53.0 & 8.83 \\

+MaPPO
& \textbf{28.84} {\scriptsize\textcolor{ForestGreen}{(+1.08)}}
& \textbf{64.4} {\scriptsize\textcolor{ForestGreen}{(+11.4)}}
& \textbf{8.84} {\scriptsize\textcolor{ForestGreen}{(+0.01)}} \\

\cline{1-4}

CPO
& 32.94 & 47.6 & 8.62 \\

+MaPPO
& \textbf{33.71} {\scriptsize\textcolor{ForestGreen}{(+0.77)}}
& \textbf{54.1} {\scriptsize\textcolor{ForestGreen}{(+6.5)}}
& \textbf{8.68} {\scriptsize\textcolor{ForestGreen}{(+0.06)}} \\

\bottomrule[1.0pt]
\end{tabular}
}

\end{wraptable}


Overall, the gains are substantial across several metrics. For example, on AlpacaEval 2.0 with Mistral-7B-Instruct, MaPPO improves DPO from $18.24$ to $30.56$, an absolute gain of $12.32$ points, and improves I-DPO from $17.11$ to $33.28$, an absolute gain of $16.17$ points. For Qwen2.5-3B-Instruct, MaPPO improves DPO from $20.16$ to $26.68$, an absolute gain of $6.52$ points, and improves I-DPO from $19.69$ to $25.99$, an absolute gain of $6.30$ points.

\begin{table}[t]
\caption{Main evaluation results on three standard benchmarks. $\uparrow$ indicates the higher the better.}
\centering
\label{table:Evaluation}
\begin{tabular}{l|l|lll}
\toprule[1.1pt]
{Model} & {Method} & {AlpacaEval 2.0 $\uparrow$} & {Arena-Hard $\uparrow$} & {MT-Bench $\uparrow$} \\
\midrule[1.1pt]
\multirow{5}{*}{Mistral-7B-Instruct} & {IT}  & {$15.35$ } &{$13.1$}&{$5.40$} \\
\cline{2-5}
{ } & {DPO} & {$18.24$} &{$14.2$}&{$6.86$} \\
{ } & {+\textbf{MaPPO}} & {$\mathbf{30.56}$ $\textcolor{ForestGreen}{+12.32}$} & {$\mathbf{18.4}$ $\textcolor{ForestGreen}{+4.2}$} & {$\mathbf{7.51}$ $\textcolor{ForestGreen}{+0.65}$} \\
\cline{2-5}
{ } & {I-DPO} & {$17.11$}  & {$14.3$}  & {$6.92$} \\
{ } & {+\textbf{MaPPO}} & {$\mathbf{33.28}$ $\textcolor{ForestGreen}{+16.17}$} &{$\mathbf{19.6}$ $\textcolor{ForestGreen}{+5.3}$} & {$\mathbf{7.59}$ $\textcolor{ForestGreen}{+0.67}$}\\
\midrule[1.1pt]
\multirow{5}{*}{Llama-3-8B-Instruct} & {IT} & {$10.85$} &{$10.2$}&{$7.52$}\\
\cline{2-5}
{ } & {DPO} & {$22.48$} &{$22.4$}&{$8.07$} \\
{ } & {+\textbf{MaPPO}} & {$\mathbf{28.37}$ $\textcolor{ForestGreen}{+5.89}$} & {$\mathbf{29.5}$ $\textcolor{ForestGreen}{+7.1}$} & {$\mathbf{8.18}$ $\textcolor{ForestGreen}{+0.11}$} \\
\cline{2-5}
{ } & {I-DPO} & {${29.47}$} &{$25.6$ }&{$8.01$}\\
{ } & {+\textbf{MaPPO}} & {$\mathbf{32.68}$ $\textcolor{ForestGreen}{+3.21}$} & {$\mathbf{31.0}$ $\textcolor{ForestGreen}{+5.4}$} & {$\mathbf{8.04}$ $\textcolor{ForestGreen}{+0.03}$} \\
\midrule[1.1pt]
\multirow{5}{*}{Qwen2.5-1.5B-Instruct} & {IT} & {11.10} &{5.0}&{7.06} \\
\cline{2-5}
{ } & {DPO} & {$18.71$} &{$11.6$}&{$7.29$} \\
{ } & {+\textbf{MaPPO}} & {$\mathbf{19.35}$ $\textcolor{ForestGreen}{+0.64}$} & {$\mathbf{15.3}$ $\textcolor{ForestGreen}{+3.7}$} & {$\mathbf{7.57}$ $\textcolor{ForestGreen}{+0.28}$} \\
\cline{2-5}
{ } & {I-DPO} & {$17.89$} &{$12.1$}&{$7.39$} \\
{ } & {+\textbf{MaPPO}} & {$\mathbf{19.84}$ $\textcolor{ForestGreen}{+1.95}$} & {$\mathbf{15.7}$ $\textcolor{ForestGreen}{+3.6}$} & {$\mathbf{7.63}$ $\textcolor{ForestGreen}{+0.24}$} \\
\midrule[1.1pt]
\multirow{5}{*}{Qwen2.5-3B-Instruct} & {IT} & {$18.91$} &{$24.0$}&{$7.92$}\\
\cline{2-5}
{ } & {DPO} & {$20.16$} &{$29.2$}&{${8.02}$} \\
{ } & {+\textbf{MaPPO}} & {$\mathbf{26.68}$ $\textcolor{ForestGreen}{+6.52}$} & {$\mathbf{35.1}$ $\textcolor{ForestGreen}{+5.9}$} & {$\mathbf{8.13}$ $\textcolor{ForestGreen}{+0.11}$} \\
\cline{2-5}
{ } & {I-DPO} & {$19.69$} &{$\mathbf{36.6}$}&{$\mathbf{8.10}$} \\
{ } & {+\textbf{MaPPO}} & {$\mathbf{25.99}$ $\textcolor{ForestGreen}{+6.30}$} & {$35.8$ $\textcolor{Apricot}{-0.8}$} & {$8.01$ $\textcolor{Apricot}{-0.09}$} \\
\midrule[1.1pt]
\multirow{5}{*}{Qwen2.5-7B-Instruct} & {IT} & {$27.03$}  & {$42.9$ }  & {$8.61$ } \\
\cline{2-5}
{ } & {DPO} & {$32.01$} &{$45.5$}&{$8.56$} \\
{ } & {+\textbf{MaPPO}} & {$\mathbf{38.24}$ $\textcolor{ForestGreen}{+6.23}$} & {$\mathbf{59.2}$ $\textcolor{ForestGreen}{+13.7}$} & {$\mathbf{8.79}$ $\textcolor{ForestGreen}{+0.23}$} \\
\cline{2-5}
{ } & {I-DPO} & {$33.80$} & {$46.9$} & {$\mathbf{8.55}$}\\
{ } & {+\textbf{MaPPO}} & {$\mathbf{39.10}$ $\textcolor{ForestGreen}{+5.30}$} & {$\mathbf{61.6}$ $\textcolor{ForestGreen}{+14.7}$} & {$8.54$ $\textcolor{Apricot}{-0.01}$}\\
\midrule[1.1pt]
\multirow{5}{*}{Qwen2.5-14B-Instruct} & {IT} & {$31.09$} &{$43.6$}&{$8.62$}\\
\cline{2-5}
{ } & {DPO} & {$39.84$} &{$52.3$}&{$8.46$} \\
{ } & {+\textbf{MaPPO}} & {$\mathbf{44.92}$ $\textcolor{ForestGreen}{+5.08}$} & {$\mathbf{59.1}$ $\textcolor{ForestGreen}{+6.8}$} & {$\mathbf{8.68}$ $\textcolor{ForestGreen}{+0.22}$} \\
\cline{2-5}
{ } & {I-DPO} & {${45.07}$} &{$54.5$} & {$8.64$}\\
{ } & {+\textbf{MaPPO}} & {$\mathbf{48.89}$ $\textcolor{ForestGreen}{+3.82}$} & {$\mathbf{62.1}$ $\textcolor{ForestGreen}{+7.6}$} & {$\mathbf{8.80}$ $\textcolor{ForestGreen}{+0.16}$} \\
\midrule[1.1pt]
\multirow{5}{*}{Qwen2.5-32B-Instruct} & {IT} & {$35.26$} &{$74.2$}&{$8.89$}\\
\cline{2-5}
{ } & {DPO} & {$46.32$} &{$81.4$}&{$8.47$} \\
{ } & {+\textbf{MaPPO}} & {$\mathbf{51.68}$ $\textcolor{ForestGreen}{+5.36}$} & {$\mathbf{87.1}$ $\textcolor{ForestGreen}{+5.7}$} & {$\mathbf{8.90}$ $\textcolor{ForestGreen}{+0.43}$} \\
\cline{2-5}
{ } & {I-DPO} & {${51.12}$} &{$86.0$ }&{$8.97$}\\
{ } & {+\textbf{MaPPO}} & {$\mathbf{58.68}$ $\textcolor{ForestGreen}{+7.56}$} & {$\mathbf{89.8}$ $\textcolor{ForestGreen}{+3.8}$} & {$\mathbf{8.99}$ $\textcolor{ForestGreen}{+0.02}$} \\
\bottomrule[1.1pt]
\end{tabular}
\end{table}

In Table~\ref{table:adaptation_main}, we present the performance improvements achieved by applying MaPPO to vanilla DPO~\citep{dpo_neurips2023} and its variants, including widely used methods such as I-DPO~\citep{dong2024rlhf}, SimPO~\citep{meng2024simpo}, IPO~\citep{pmlr-v238-gheshlaghi-azar24a}, and CPO~\citep{xu2024contrastive}. All experiments are conducted using Qwen2.5-7B-Instruct. The overall results further demonstrate the broad compatibility of MaPPO and consistently validate its effectiveness. More detailed discussions are provided in Appendix~\ref{app:adapt}.


\subsection{Other Results}

Additional experimental results are provided in Appendix \ref{app:supp_results} for further evaluation of MaPPO.

\paragraph{Adaptation to DPO Variants.}
In Appendix \ref{app:adapt}, we provide a more detailed discussions on the improvement of the vanilla DPO~\citep{dpo_neurips2023} and its variants with MaPPO, including widely used I-DPO~\citep{dong2024rlhf}, SimPO~\citep{meng2024simpo}, IPO~\citep{pmlr-v238-gheshlaghi-azar24a}, and CPO~\citep{xu2024contrastive}. The corresponding equations are provided in Appendix \ref{app:adapt_other_po}.

\paragraph{Academic Benchmarks.}
In Appendix \ref{app:Academic}, we test the performance on six widely used academic benchmarks, including IFEval, GPQA, MMLU, HellaSwag, TruthfulQA and GSM8K. It shows that the gains observed on MT-Bench, AlpacaEval 2.0, and Arena-Hard generalize to a broader set of evaluation tasks.

\paragraph{Broader Evaluation.}
In Appendix \ref{app:Broader_Evaluation}, we provide human evaluation, and the model performance with a direct head-to-head comparison between the baseline and MaPPO-enhanced modes.

\paragraph{Reward Signals.}
To better understand the role of RMs, in Appendix \ref{app:Ablation_RMs}, we conduct an ablation study by integrating different open-source RMs into our framework, highlighting the robustness of our approach under varying prior signals. We justify the equivalence of prior knowledge design and RMs in Appendix \ref{app:rm_equal}.

\paragraph{Compatibility.}
Finally, in Appendix \ref{app:Comparison}, we compare our approach with complementary SFT-B, illustrating that MaPPO provides additive benefits and can be effectively combined with SFT-B.

\section{Related Work}
\label{Sec:short_Related_Work}

Here we discuss the most relevant prior work and leave a broader related work to Appendix \ref{Sec:Related_Work}.

Recent preference-optimization research has largely focused on improving the efficiency and stability of DPO-style alignment methods. DPO~\citep{dpo_neurips2023} casts preference learning as maximum-likelihood estimation under the Bradley--Terry model~\citep{bradley1952rank}, and follow-up variants such as IPO~\citep{pmlr-v238-gheshlaghi-azar24a}, CPO~\citep{xu2024contrastive}, KTO~\citep{ethayarajh2024model}, ORPO~\citep{hong-etal-2024-orpo}, R-DPO~\citep{park2024disentangling}, SimPO~\citep{meng2024simpo}, Weighted-Reward PO~\citep{yang2025weightedreward}, and RainbowPO~\citep{zhao2025rainbowpo} extend this framework through alternative objectives, regularization, or weighting strategies. At the same time, several works have identified a key limitation of the MLE formulation: It can reduce the absolute likelihood of both preferred and rejected responses, producing confidence degeneration and the \textit{squeezing effect}~\citep{pal2024smaug, rafailov2024r, tajwar2024preference, ren2024learning}. Nemotron-4~\citep{adler2024nemotron} adds a reward-margin constraint, but it does not provide a general theoretical account and is less flexible for adaptation to other DPO variants. In contrast, MaPPO injects prior reward knowledge directly into the objective, preserving DPO-style efficiency while improving calibration and mitigating these instability issues.

\section{Discussions}
\label{Sec:Discussions}

\subsection{Limitations and Future Work}


\begin{enumerate}
\item Our results indicate that larger models consistently perform better with the MaPPO method. Future work with more computing resources could explore applying the proposed training pipeline to models larger than 32B parameters.
\item The prior knowledge function design relies on experts' domain knowledge. We give a reasonable and general design. Future works could explore different designs in specific domains.
\item Human evaluation is limited to 1,000 prompts and one model family. If resources permit, human evaluation can be scaled for better judgment of alignment.
\end{enumerate}

\subsection{Broader Impacts}
\label{app:Broader_Impacts}

MaPPO aligns language models by incorporating prior reward knowledge into preference optimization, leading to better-calibrated and more robust outputs. 
While beneficial, it relies on reward models that may encode biases or misrepresent human values, potentially reinforcing harmful patterns. 
Its use in persuasive or deceptive applications also poses potential impacts. 
To mitigate these impacts, we encourage careful curation and auditing of reward models, broader involvement in defining reward signals, and transparency in how preference optimization frameworks, such as MaPPO, are applied in real-world AI systems.

\subsection{Conclusion}

We propose MaPPO, a general and principled framework for preference optimization that incorporates prior knowledge into the optimization objective. 
By extending the Maximum Likelihood Estimation (MLE)-based Preference Optimization (PO) approach to a Maximum a Posteriori (MaP) formulation, MaPPO effectively mitigates confidence degeneration and provides a calibrated training signal. 
Our method requires no additional hyperparameters, supports both offline and online settings, and can be seamlessly integrated into existing Direct Preference Optimization (DPO) variants, including widely used SimPO, IPO, and CPO.
Without sacrificing efficiency, extensive empirical results demonstrate that MaPPO consistently improves alignment performance on different model series (Qwen, Mistral, and Llama), and on scaling to different model sizes (1.5B, 3B, 7B, 8B, 14B, and 32B) across three standard benchmarks, including MT-Bench, AlpacaEval 2.0, and Arena-Hard.
We further evaluate the performance on six academic benchmarks, which shows that MaPPO maintains the performance in various dimensions.


\section*{Acknowledgments}
We thank all reviewers for their invaluable feedback that helped us significantly strengthen the work. C. Brinton acknowledges support from the National Science Foundation (NSF) under grant CNS-2146171, the Office of Naval Research (ONR) under grant N00014-22-1-2305, and the Air Force Office of Scientific Research (AFOSR) under grant FA9550-24-1-0083.

{
\bibliographystyle{abbrvnat}
\bibliography{ref}
}

\clearpage
\newpage
\appendix

\begin{center}
    {\bf\Large Appendix}
\end{center}

\startcontents[sections]
\printcontents[sections]{l}{1}{\setcounter{tocdepth}{4}}

\clearpage
\newpage
\section{Supplementary Related Work}
\label{Sec:Related_Work}

\subsection{Direct Preference Optimization and its Variants.}

Driven by the complexity of online RLHF algorithms~\citep{santacroce2023efficient, zheng2023secrets}, recent research has pivoted toward efficient offline preference optimization.
Direct Preference Optimization (DPO)~\citep{dpo_neurips2023} frames preference alignment as maximum-likelihood estimation (MLE) under the Bradley-Terry (BT) model~\citep{bradley1952rank}, while IPO~\citep{pmlr-v238-gheshlaghi-azar24a} generalizes this framework without the pointwise‑reward assumption. Further, CPO~\citep{xu2024contrastive} jointly optimizes the sequence likelihood and a contrastive reward to perform supervised fine-tuning (SFT) and alignment in one pass. 
KTO~\citep{ethayarajh2024model} extends the paradigm to single‑response feedback via prospect theoretic utility. 
Recent DPO variants, ORPO~\citep{hong-etal-2024-orpo}, R-DPO~\citep{park2024disentangling}, and SimPO~\citep{meng2024simpo}, further push performance by discarding the reference model or regularizing response length.
\cite{yang2025weightedreward} further introduces a weight hyperparameter to balance the influence of preference pairs from different policies. 
\cite{zhao2025rainbowpo} then aims to combine the previous PO methods into one cohesive objective. 
However, all DPO-style variants rely on MLE in the training process, which oversimplifies the tuning of preferred and unpreferred responses as a binary classification problem.

Beyond preference optimization, recent work has also improved the efficiency, scalability, and trustworthiness of ML/LLM pipelines \cite{su2024fastmatch,su2026sair,su2026trustweave}. 
More broadly, recent studies have examined LLM fine-tuning dynamics~\cite{zeng2025lensllm} and hallucinations~\cite{zeng2026halluguard}.
Related studies have explored multimodal content understanding and governance through process supervision, domain-adaptive pretraining, and multimodal data expansion \cite{wu2026injection, liu-etal-2026-ips, wang-etal-2025-reasoning-enhanced, 10.1145/3711896.3737195}.
Related post-training research has also studied machine unlearning in recommender systems, including multi-objective preference-guided approaches, attribute unlearning, and broader formulations of recommendation unlearning \cite{li2025multi,chen2024post,11270220}.
Recent works has also investigated objective design under severe class
imbalance~\cite{sun2025objective} and temporal consistency
constraints under non-stationary conditions~\cite{lei2026multi}.

\subsection{Confidence Degeneration in DPO.}

\citet{pal2024smaug} and \citet{tajwar2024preference} show analytically and empirically that the expected DPO gradient often decreases the log-likelihood of the preferred response $\mathbf{y}_w$ instead of increasing it, leading to a simultaneous shrinkage of both responses.
\citet{rafailov2024r} observe the same trend, attributing the drop to the expected log ratio between the optimized and reference models. By showing that this is equivalent to the non-negative KL divergence, they conclude that DPO training inevitably lowers the likelihood of the chosen response.
Nemotron-4 \citep{adler2024nemotron} constraints the DPO loss with the reward margin. However, no motivation or theoretical perspective is given. Moreover, it cannot be used in the online setting and is unable to enhance the other DPO variants.
More recent analyses of the learning dynamics in \cite{ren2024learning} have identified a phenomenon termed the \textit{squeezing effect}, whereby DPO training aggressively drains probability mass from all responses except the most confident one, $\mathbf{y}^*= \arg\max_{i \in [V ]\setminus {\mathbf{y}_l}} \pi_\theta (\mathbf{y} = i)$, consequently funneling this mass towards $\mathbf{y}^*$. 
Our method utilizes prior knowledge to soften the downward pressure on the rejected response $\mathbf{y}_l$, it markedly mitigates the squeezing effect.

\section{Supplementary Concepts}
\label{app:Supplementary_Concepts}

\subsection{Iterative DPO Illustration}
\label{app:Illustration}

\begin{figure}[ht]
    \centering
    \includegraphics[width=1\linewidth]{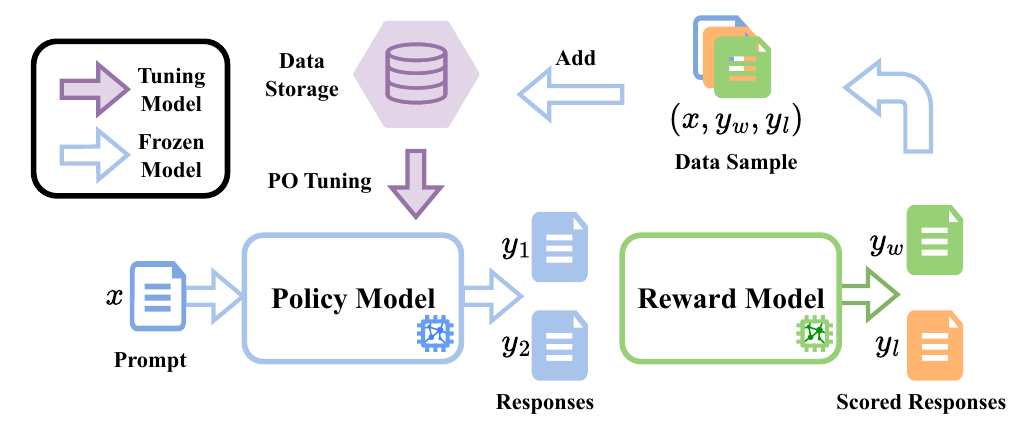}
    \caption{Illustration of the iterative MaPPO pipeline in each iteration $k$.}
    \label{fig:i_mappo}
\end{figure}

In Figure \ref{fig:i_mappo}, we illustrate the iterative MaPPO pipeline.

\subsection{Adaptation to Other PO Methods}
\label{app:adapt_other_po}

We have shown MaPPO in the offline and online DPO settings. As this Maximum a Posteriori (MaP) method is generally suitable for all DPO variants, we show how MaPPO modifies other DPO variants in this subsection.

Simply replace the MLE part in preference optimization with MaP, and follow the same derivation in Section \ref{Sec:mappo_loss}. Most DPO variants, as long as MLE is used in the original methods, can be modified with MaPPO as a plugin. We show some widely adopted methods as examples here, including SimPO, IPO, and CPO. 

First, in SimPO \citep{meng2024simpo}, with the length control penalty, the loss function is given as
\begin{equation}
\label{eq:simpo}
\begin{aligned}
    \mathcal{L}_{\rm SimPO}(\theta) ~=~ 
    \mathop{\mathbb{E}}_{( \mathbf{y}_{w}, \mathbf{y}_{l}, \mathbf{x})\sim\mathcal{D}} \Big[ -\log \sigma \Big( \frac{\beta}{|\mathbf{y}_{w}|} \log {\pi_{\theta}(\mathbf{y}_{w}|\mathbf{x})} - \frac{\beta}{|\mathbf{y}_{l}|} \log {\pi_{\theta}(\mathbf{y}_{l}|\mathbf{x})} - \gamma \Big)\Big],
\end{aligned}
\end{equation}
where $\gamma$ is a constant hyperparameter, $|\mathbf{y}_{w}|$ and $|\mathbf{y}_{l}|$ denote the lengths of $\mathbf{y}_{w}$ and $\mathbf{y}_{l}$, respectively.

With the MaPPO plugin, the loss function is modified as 
\begin{equation}
\label{eq:simpo_mappo}
\begin{aligned}
    \mathcal{L}_{\rm SimPO+}(\theta) ~=~ 
    \mathop{\mathbb{E}}_{( \mathbf{y}_{w}, \mathbf{y}_{l}, \mathbf{x})\sim\mathcal{D}} \Big[ -\log \sigma \Big( \frac{\beta}{|\mathbf{y}_{w}|} \log {\pi_{\theta}(\mathbf{y}_{w}|\mathbf{x})} - \Delta_{r} \frac{\beta}{|\mathbf{y}_{l}|} \log {\pi_{\theta}(\mathbf{y}_{l}|\mathbf{x})} - \gamma \Big)\Big].
\end{aligned}
\end{equation}

In IPO \citep{azar2024general}, the original loss function is
\begin{equation}
\label{eq:IPO}
\begin{aligned}
    \mathcal{L}_{\rm IPO}(\theta) ~=~\mathop{\mathbb{E}}_{( \mathbf{y}_{w}, \mathbf{y}_{l}, \mathbf{x})\sim\mathcal{D}} \Big[ \Big( \log \frac{\pi_{\theta}(\mathbf{y}_{w}|\mathbf{x})}{\pi_{\rm ref}(\mathbf{y}_{w}|\mathbf{x})} - \log \frac{\pi_{\theta}(\mathbf{y}_{l}|\mathbf{x})}{\pi_{\rm ref}(\mathbf{y}_{l}|\mathbf{x})} - \frac{1}{2\beta} \Big)^{2} \Big].
\end{aligned}
\end{equation}
With the MaPPO plugin, the loss function is modified as 
\begin{equation}
\label{eq:IPO_mappo}
\begin{aligned}
    \mathcal{L}_{\rm IPO+}(\theta) ~=~ 
    \mathop{\mathbb{E}}_{( \mathbf{y}_{w}, \mathbf{y}_{l}, \mathbf{x})\sim\mathcal{D}} \Big[ \Big( \log \frac{\pi_{\theta}(\mathbf{y}_{w}|\mathbf{x})}{\pi_{\rm ref}(\mathbf{y}_{w}|\mathbf{x})} - \Delta_{r} \log \frac{\pi_{\theta}(\mathbf{y}_{l}|\mathbf{x})}{\pi_{\rm ref}(\mathbf{y}_{l}|\mathbf{x})} - \frac{1}{2\beta} \Big)^{2} \Big].
\end{aligned}
\end{equation}

In CPO \citep{xu2024contrastive}, the original loss is 
\begin{equation}
\label{eq:cpo}
\begin{aligned}
    \mathcal{L}_{\rm CPO}(\theta) ~=~ 
    \mathop{\mathbb{E}}_{( \mathbf{y}_{w}, \mathbf{y}_{l}, \mathbf{x})\sim\mathcal{D}} \Big[ -\log \sigma \Big( \beta \log {\pi_{\theta}(\mathbf{y}_{w}|\mathbf{x})} - \beta \log {\pi_{\theta}(\mathbf{y}_{l}|\mathbf{x})} \Big) - \lambda \log {\pi_{\theta}(\mathbf{y}_{w}|\mathbf{x})} \Big],
\end{aligned}
\end{equation}
where $\lambda$ is a constant hyperparameter.

With the MaPPO plugin, the CPO loss is modified as 
\begin{equation}
\label{eq:CPO_mappo}
\begin{aligned}
    \mathcal{L}_{\rm CPO+}(\theta) ~=~ 
    \mathop{\mathbb{E}}_{( \mathbf{y}_{w}, \mathbf{y}_{l}, \mathbf{x})\sim\mathcal{D}} \Big[ -\log \sigma \Big( \beta \log {\pi_{\theta}(\mathbf{y}_{w}|\mathbf{x})} - \beta \Delta_{r} \log {\pi_{\theta}(\mathbf{y}_{l}|\mathbf{x})} \Big) - \lambda \log {\pi_{\theta}(\mathbf{y}_{w}|\mathbf{x})} \Big].
\end{aligned}
\end{equation}

To verify the effectiveness of this MaPPO plugin, we show the experimental results with the improvement of these DPO variants in Appendix \ref{app:adapt}.

\textbf{Remark.}
With our MaPPO plugin, \textbf{no} additional hyperparameter and computational cost is introduced in all DPO variants.

\newpage
\section{Theoretical Results}
\label{app:Theoretical_Results}

\subsection{The Equivalence of Prior Signals}
\label{app:rm_equal}

In this subsection, we mathematically formalize the following equivalence. 
\begin{proposition}
Equivalence between a class of prior designs and transformed reward models.
\end{proposition}

For a preference pair $(\mathbf{x}, \mathbf{y}_w, \mathbf{y}_l)$, define
\[
A_\theta := \beta \log \frac{\pi_\theta(\mathbf{y}_w \mid \mathbf{x})}{\pi_{\mathrm{ref}}(\mathbf{y}_w \mid \mathbf{x})},
\qquad
B_\theta := \beta \log \frac{\pi_\theta(\mathbf{y}_l \mid \mathbf{x})}{\pi_{\mathrm{ref}}(\mathbf{y}_l \mid \mathbf{x})}.
\]

The standard MaPPO loss is
\[
\ell_{\mathrm{MaPPO}}(\theta)
=
-\log \sigma \!\left(A_\theta - \Delta r \, B_\theta\right),
\qquad
\Delta r := r(\mathbf{y}_w, \mathbf{x}) - r(\mathbf{y}_l, \mathbf{x}).
\]

Now consider a generalized prior design whose only effect is to replace $\Delta_{r}$ by a scalar coefficient
\[
\alpha(\mathbf{x}, \mathbf{y}_w, \mathbf{y}_l) = g \big(r(\mathbf{y}_w, \mathbf{x}), r(\mathbf{y}_l, \mathbf{y})\big).
\]
The resulting loss becomes
\[
\ell_g(\theta)
=
-\log \sigma \!\left(A_\theta - \alpha(\mathbf{x}, \mathbf{y}_w, \mathbf{y}_l) \, B_\theta\right).
\]

Then the following statements hold.

\textit{(1) Exact equivalence at the level of the effective calibration signal.}

Define
\[
\widetilde{\Delta}(\mathbf{x}, \mathbf{y}_w, \mathbf{y}_l) := g \big(r(\mathbf{y}_w, \mathbf{x}), r(\mathbf{y}_l, \mathbf{x})\big).
\]
Then
\[
\ell_g(\theta)
=
-\log \sigma \!\left(A_\theta - \widetilde{\Delta}(\mathbf{x}, \mathbf{y}_w, \mathbf{y}_l) \, B_\theta\right),
\]
which is exactly the standard MaPPO objective with effective calibration coefficient $\widetilde{\Delta}$. Hence, the two objectives are pointwise identical for every $\theta$, and therefore they also have identical gradients:
\[
\nabla_\theta \ell_g(\theta)
=
\nabla_\theta \ell_{\mathrm{MaPPO}}(\theta;\widetilde{\Delta}).
\]
Thus, under the same optimizer, initialization, and minibatches, the optimization trajectory is the same.

\textit{(2) Exact equivalence to a different reward model under a mild structural condition.}

Assume that the alternative prior design is difference-separable, i.e.,
\[
g(a,b) = T(a) - T(b)
\]
for some scalar function $T$. Define a transformed reward model
\[
\widetilde r(\mathbf{y}, \mathbf{x}) := T(r(\mathbf{y}, \mathbf{x})).
\]
Then
\[
\widetilde r(\mathbf{y}_w, \mathbf{x}) - \widetilde r(\mathbf{y}_l, \mathbf{x})
=
T\Big(r(\mathbf{y}_w, \mathbf{x})) - T(r(\mathbf{y}_l, \mathbf{x})\Big)
=
g\Big(r(\mathbf{y}_w, \mathbf{x}), r(\mathbf{y}_l, \mathbf{x})\Big).
\]
Substituting this into the loss gives
\[
\ell_g(\theta)
=
-\log \sigma \!\left(
A_\theta
-
\big(\widetilde r(\mathbf{y}_w, \mathbf{x}) - \widetilde r(\mathbf{y}_l, \mathbf{x})\big) B_\theta
\right),
\]
which is exactly the standard MaPPO loss with the same prior form and the transformed reward model $\widetilde{r}$.

Part (1) follows immediately by defining $\widetilde{\Delta}(\mathbf{x}, \mathbf{y}_w, \mathbf{y}_l) = g(r(\mathbf{y}_w, \mathbf{x}), r(\mathbf{y}_l, \mathbf{x}))$ and substituting it into the loss. Since the two losses are equal for every parameter value $\theta$, their gradients with respect to $\theta$ are also equal.

For part (2), if $g$ is difference-separable, define $\widetilde{r}(\mathbf{y}, \mathbf{x}) = T\big(r(\mathbf{y}, \mathbf{x})]\big)$. Then
\[
\widetilde{r}(\mathbf{y}_w, \mathbf{x})-\widetilde{r}(\mathbf{y}_l, \mathbf{x}) = g\Big(r(\mathbf{y}_w, \mathbf{x}), r(\mathbf{y}_l, \mathbf{x})\Big),
\]
so the generalized-prior loss can be rewritten exactly as the standard MaPPO loss with reward model $\widetilde{r}$. This proves the equivalence. $\square$

We emphasize that this equivalence holds for the broad and practically relevant class of prior designs that enter MaPPO only through a scalar calibration coefficient, and it becomes an exact reward-model equivalence when that coefficient is difference-separable.

\subsection{Stationary Convergence of MaPPO}
\label{app:theo_converge}

Recall that 
\begin{equation}
\label{eq:mappo_loss_u}
\begin{aligned}
    \mathcal{L}_{\rm MaP}(\theta) ~=~& 
    \mathop{\mathbb{E}}_{( \mathbf{y}_{w}, \mathbf{y}_{l}, \mathbf{x})\sim\mathcal{D}} \Big[ -\log \sigma \Big( \beta\log \frac{\pi_{\theta}(\mathbf{y}_{w}|\mathbf{x})}{\pi_{\rm ref}(\mathbf{y}_{w}|\mathbf{x})} - \Delta_{r} \beta\log \frac{\pi_{\theta}(\mathbf{y}_{l}|\mathbf{x})}{\pi_{\rm ref}(\mathbf{y}_{l}|\mathbf{x})} \Big)\Big] \\
    ~=~& \mathop{\mathbb{E}}_{( \mathbf{y}_{w}, \mathbf{y}_{l}, \mathbf{x})\sim\mathcal{D}} \Big[ -\log \sigma (u) \Big].
\end{aligned}
\end{equation}

At the first-order stationary point (FOSP), the gradient of the loss with respect to $\theta$ becomes $0$. Thus, we have 
\begin{equation}
\label{eq:mappo_grad_0}
\begin{aligned}
    \nabla \mathcal{L}_{\rm MaP}(\theta) ~=~&
    \mathop{\mathbb{E}}_{( \mathbf{y}_{w}, \mathbf{y}_{l}, \mathbf{x})\sim\mathcal{D}} \Big[ -\beta (1-\sigma(u)) \Big( \nabla \log {\pi_{\theta}(\mathbf{y}_{w}|\mathbf{x})} - \Delta_{r} \nabla \log {\pi_{\theta}(\mathbf{y}_{l}|\mathbf{x})} \Big)\Big] = 0.
\end{aligned}
\end{equation}

Let the optimal policy be $\pi_{\star}$. The above equation holds when 
\begin{equation}
\label{eq:mappo_grad_inside_0}
\begin{aligned}
\nabla \log {\pi_{\star}(\mathbf{y}_{w}|\mathbf{x})} - \Delta_{r} \nabla \log {\pi_{\star}(\mathbf{y}_{l}|\mathbf{x})} ~=~ 0.
\end{aligned}
\end{equation}

Thus, the optimal policy achieves
\begin{equation}
\label{eq:mappo_opt}
\begin{aligned}
\log {\pi_{\star}(\mathbf{y}_{w}|\mathbf{x})} ~=~ \Delta_{r} \log {\pi_{\star}(\mathbf{y}_{l}|\mathbf{x})} + c,
\end{aligned}
\end{equation}
where $c$ is a scaling constant determined by the initialization. Thus, the model learns a stable log-linear relationship between preferred and less-preferred responses, scaled by the prior reward gap.

In DPO, the optimal policy at the FOSP is
\begin{equation}
\label{eq:dpo_opt}
\begin{aligned}
\log {\pi_{\star}(\mathbf{y}_{w}|\mathbf{x})} ~=~ \log {\pi_{\star}(\mathbf{y}_{l}|\mathbf{x})} + c.
\end{aligned}
\end{equation}
DPO converges to maximizing the log-odds between $\mathbf{y}_{w}$ and $\mathbf{y}_{l}$, but no inherent bound on the preference gap, which can lead to confidence degeneration. 
As training progresses, DPO may tend to decrease the likelihood of both $\mathbf{y}_{w}$ and $\mathbf{y}_{l}$ (the \emph{squeezing effect}), because there is no constraint on absolute probabilities -- only the relative gap matters.

MaPPO prevents overconfidence and instability by grounding optimization in the reward-based prior. 
The FOSP of MaPPO guarantees a bounded, calibrated log-probability ratio between $\mathbf{y}_{w}$ and $\mathbf{y}_{l}$. 
Naturally limits the \emph{squeezing effect} by scaling the impact of $\mathbf{y}_{l}$ via $\Delta_{r}$.
Therefore, MaPPO is theoretically more stable, especially for near-tie preference pairs and in large models where DPO can exacerbate miscalibration.

\subsection{Lipschitz Stability}
\label{app:Lipschitz}

First, we list the standard assumptions for the analysis.
\begin{assumption}{\ }
\label{assum:bound}
\begin{enumerate}
\item The score function is Lipschitz continuous as $\left\|\nabla \log \pi_\theta (\mathbf{y} | \mathbf{x}) - \nabla \log \pi_{\theta'} (\mathbf{y} | \mathbf{x}) \right\| \leq M_{g} \|\theta - \theta'\|$.
\end{enumerate}
\end{assumption}

Let the gradient operator be defined as
\begin{equation}
\label{eq:lip_mappo}
\begin{aligned}
\tau_{\theta} ~\coloneqq~  \nabla \mathcal{L}_{\rm MaP}(\theta).
\end{aligned}
\end{equation}

Then, the gradient operator $\tau$ is Lipschitz continuous with
\begin{equation}
\label{eq:lip_bound_mappo}
\begin{aligned}
\| \tau_{\theta} - \tau_{\theta'} \| \le L_{\rm MaP} \| \theta - \theta' \|,
\end{aligned}
\end{equation}
where $L_{\rm MaP} = \beta (1 - \sigma(u)) (1 + \Delta_{r}) M_{g} < \beta (1 + \Delta_{r}) M_{g}$.

\begin{proof}
We have
\begin{equation}
\begin{aligned}
\tau_{\theta} ~=~ -\beta (1-\sigma(u)) \Big( \nabla \log {\pi_{\theta}(\mathbf{y}_{w}|\mathbf{x})} - \Delta_{r} \nabla \log {\pi_{\theta}(\mathbf{y}_{l}|\mathbf{x})} \Big).
\end{aligned}
\end{equation}

The norm of the gradient difference is 
\begin{equation}
\begin{aligned}
\| \tau_{\theta} - \tau_{\theta'} \| ~\leq&~ \beta (1-\sigma(u)) \cdot \| \nabla \log {\pi_{\theta}(\mathbf{y}_{w}|\mathbf{x})} - \nabla \log {\pi_{\theta'}(\mathbf{y}_{w}|\mathbf{x})} \| \\
&+ \beta (1-\sigma(u)) \Delta_{r} \cdot \| \nabla \log {\pi_{\theta}(\mathbf{y}_{l}|\mathbf{x})} - \nabla \log {\pi_{\theta'}(\mathbf{y}_{l}|\mathbf{x})} \|.
\end{aligned}
\end{equation}

We have that $\sigma(u)$ is Lipschitz with constant $\frac{1}{4}$. Combining the $M_{g}$ Lipschitz of the score function $\nabla \log \pi_\theta (\mathbf{y} | \mathbf{x})$, we have
\begin{equation}
\begin{aligned}
\| \tau_{\theta} - \tau_{\theta'} \| \leq \beta (1 - \sigma(u)) (1 + \Delta_{r}) M_{g} \| \theta - \theta' \| ~\coloneqq~ L_{\rm MaP} \| \theta - \theta' \|.
\end{aligned}
\end{equation}
\end{proof}

In contrast, in DPO, the gradient is Lipschitz continuous as
\begin{equation}
\label{eq:lip_bound_dpo}
\begin{aligned}
\| \tau_{\theta} - \tau_{\theta'} \| \le L_{\rm DPO} \| \theta - \theta' \|,
\end{aligned}
\end{equation}
where $L_{\rm DPO} = 2\beta M_{g} > 2\beta (1 - \sigma(u)) M_{g} $. The Lipschitz constant of the gradients in DPO is larger than that in MaPPO, which shows the gradients have less stability in DPO.

With a formal upper bound on MaPPO's gradient variation, we show a clear theoretical justification for why MaPPO is less prone to instability and exploding confidence gaps, compared to DPO.
\clearpage
\newpage
\section{Supplementary Experiments}
\label{app:Experiments}

\subsection{Supplementary Experimental Settings}
\label{sec:app_exp_setting}

\subsubsection{Hyperparameter settings.} 
We follow the standard settings and list the hyperparameter details in the training process of MaPPO in Table \ref{table:hyper}. 
We keep the hyperparameter settings for different model series and model sizes, including Qwen2.5-\{1.5B, 3B, 7B\}-Instruct, Mistral-7B-Instruct, and Llama-3-8B-Instruct models. 

\begin{table}[!htbp]
\caption{Hyperparameter settings in MaPPO.}
\centering
\label{table:hyper}
\begin{tabular}{l|lc}
\toprule[1.1pt]
{Hyperparameter} & {Value} & {} \\
\midrule[1.1pt]
{global batch size} & {$128$} & { } \\
{learning rate $\eta$} & {$5\times 10^{-7}$} & { } \\
{warmup steps} & {$100$} & { } \\
{weight decay} & {$0.01$} & { } \\
{optimizer} & {AdamW} & { } \\
{KL weight $\beta$} & {$0.1$} & { } \\
{number of responses $n$} & {$8$} & { } \\
{temperature $T$} & {$1.0$} & { } \\
{precision} & {${\tt bfloat16}$} & { } \\
\bottomrule[1.1pt]
\end{tabular}
\end{table}

\begin{table}[!htbp]
\caption{Hyperparameter settings in DPO Variants.}
\centering
\label{table:hyper_variants}
\begin{tabular}{l|lc}
\toprule[1.1pt]
{Hyperparameter} & {Value} & {} \\
\midrule[1.1pt]
{SimPO: $\gamma$} & {$1$} & { } \\
{IPO: $\beta$} & {$0.1$} & { } \\
{CPO: $\lambda$} & {$0.2$} & { } \\
\bottomrule[1.1pt]
\end{tabular}
\end{table}

In Table \ref{table:hyper_variants}, we list the extra hyperparameters in the reproduction of the DPO variants. 
The other hyperparameter settings keep the same in Table \ref{table:hyper}. 
Notably, we choose nearly the best hyperparameters for the other methods, and our reproduction achieves higher performances than the original or other reproduction reports on some benchmarks, \textit{e.g.}, SimPO on Arena-Hard. 

\subsubsection{Computing Resources.} 
All tasks are trained and evaluated on a platform with $8$ NVIDIA H100 GPUs on each node, and $80$ GB of memory for each GPU. 
Each training task requires between $4$ and $40$ hours to execute, depending on the size of the model.

\subsection{Supplementary Results}
\label{app:supp_results}

\subsubsection{Adaptation to DPO Variants}
\label{app:adapt}

In Table \ref{table:adaptation}, we show the improvement of the vanilla DPO~\citep{dpo_neurips2023} and its variants with MaPPO, including widely used I-DPO~\citep{dong2024rlhf}, SimPO~\citep{meng2024simpo}, IPO~\citep{pmlr-v238-gheshlaghi-azar24a}, and CPO~\citep{xu2024contrastive}. 
We list the hyperparameter settings in the DPO variants in Appendix \ref{app:Experiments}. 
Their loss functions with MaPPO adaptation are shown in Section \ref{sec:online_mappo} and Appendix \ref{app:adapt_other_po}. 
For all DPO variants, no additional hyperparameter is needed from the MaPPO plugin. For the model in evaluation, we keep Qwen2.5-7B-Instruct as the default model. 

In general, MaPPO consistently improves all DPO variants with the MLE design on all three benchmarks.
Although it drops a little on MT-Bench in some methods, the original approach has essentially saturated at the achievable score on MT-Bench, which barely reflects the improvement with a variance in evaluation.
The overall consistent improvements observed across DPO variants after applying the MaPPO plugin underscore its flexibility and generality in enhancing preference optimization methods. 
Notably, MaPPO effectively complements both simple and complex variants without requiring architectural modifications or hyperparameter tuning. 
For instance, SimPO benefits from the MaPPO adjustment by further balancing the length-controlled optimization with better calibration of confidence scores, while IPO and CPO experience gains due to MaPPO’s capacity to regularize reward signals with prior knowledge, mitigating overfitting to pairwise preferences. 
The improvements span diverse evaluation metrics and benchmarks, demonstrating that MaPPO’s reward-aware calibration systematically addresses the shortcomings of MLE-based objectives inherent in existing variants. 
This indicates that MaPPO is not merely a tweak, but a general principle that can be seamlessly integrated into the PO pipelines to achieve more reliable alignment results.

\begin{table}[!ht]
\caption{Evaluation results of DPO variants with a MaPPO plugin. $\uparrow$ indicates the higher the better. The results confirm the broad applicability of MaPPO across existing methods.}
\centering
\label{table:adaptation}
\begin{tabular}{l|lll}
\toprule[1.1pt]
{Method} & {AlpacaEval 2.0 $\uparrow$} & {Arena-Hard $\uparrow$} & {MT-Bench $\uparrow$} \\
\midrule[1.1pt]
{DPO} & {$32.01$} &{$45.5$}&{$8.56$} \\
{+{MaPPO}} & {$\mathbf{38.24}$ $\textcolor{ForestGreen}{+6.23}$} & {$\mathbf{59.2}$ $\textcolor{ForestGreen}{+13.7}$} & {$\mathbf{8.79}$ $\textcolor{ForestGreen}{+0.23}$} \\
\cline{1-4}
{I-DPO} & {$33.80$} &{$46.9$}&{$\mathbf{8.55}$} \\
{+{MaPPO}} & {$\mathbf{39.10}$ $\textcolor{ForestGreen}{+5.30}$} & {$\mathbf{61.6}$ $\textcolor{ForestGreen}{+14.7}$} & {$8.54$ $\textcolor{Apricot}{-0.01}$} \\
\cline{1-4}
{SimPO} & {$25.15$} &{$64.2$}&{$\mathbf{9.02}$} \\
{+{MaPPO}} & {$\mathbf{32.75}$ $\textcolor{ForestGreen}{+7.60}$} & {$\mathbf{69.5}$ $\textcolor{ForestGreen}{+5.3}$} & {${8.94}$ $\textcolor{Apricot}{-0.08}$} \\
\cline{1-4}
{IPO} & {$27.76$} &{$53.0$}&{$8.83$} \\
{+{MaPPO}} & {$\mathbf{28.84}$ $\textcolor{ForestGreen}{+1.08}$} & {$\mathbf{64.4}$ $\textcolor{ForestGreen}{+11.4}$} & {$\mathbf{8.84}$ $\textcolor{ForestGreen}{+0.01}$} \\
\cline{1-4}
{CPO} & {$32.94$} &{$47.6$}&{8.62} \\
{+{MaPPO}} & {$\mathbf{33.71}$ $\textcolor{ForestGreen}{+0.77}$} & {$\mathbf{54.1}$ $\textcolor{ForestGreen}{+6.5}$} & {$\mathbf{8.68}$ $\textcolor{ForestGreen}{+0.06}$} \\
\bottomrule[1.1pt]
\end{tabular}
\vspace{-4mm}
\end{table}

\subsubsection{Results on Academic Benchmarks}
\label{app:Academic}

It is widely observed that alignment impairs models' performance on calibration, reasoning, and accuracy \citep{ouyang2022training, LinL0DLZP00ZDPZ24, zhang2025iterative}, which is also known as the alignment tax. As a result, it is also needed to assess the model’s performance using more academic benchmarks. In this subsection, we investigate whether the several methods for alignment with human preference could sacrifice the general model performance.

\begin{table}[!ht]
\caption{Evaluation results on six academic benchmarks with Llama-3-8B-Instruct model.}
\centering
\label{table:academic}
\begin{tabular}{l|cccccc}
\toprule[1.1pt]
 {Method} & {IFEval $\uparrow$} & {GPQA $\uparrow$} & {MMLU $\uparrow$} & {HellaSwag $\uparrow$} & {TruthfulQA $\uparrow$} & {GSM8K $\uparrow$} \\
\midrule[1.1pt]
{IT} & {$70.4$} & {$30.2$} & {$62.4$} & {$78.6$} & {$53.7$} & {$73.4$} \\
\hline
{DPO} & {$77.0$} & {$27.5$} & {$62.7$} & {$79.5$} & {$51.5$} & {$75.5$}  \\
{+\textbf{MaPPO}} & {$\mathbf{82.0}$} & {$\mathbf{29.5}$} & {$\mathbf{63.2}$} & {$\mathbf{80.1}$} & {$\mathbf{58.2}$} 
& {$\mathbf{79.5}$} \\
\hline
{I-DPO} & {$74.6$} & {$\mathbf{29.8}$} & {$63.1$} & {$80.5$} & {$60.7$} & {$81.3$} \\
{+\textbf{MaPPO}} & $\mathbf{76.4}$ & ${28.8}$ & $\mathbf{63.5}$ & $\mathbf{80.7}$ & $\mathbf{63.7}$ & $\mathbf{82.4}$ \\
\bottomrule[1.1pt]
\end{tabular}
\end{table}
\begin{table}[!ht]
\caption{Evaluation results on six academic benchmarks with Qwen2.5-7B-Instruct model.}
\centering
\label{table:academic_qwen_7b}
\begin{tabular}{l|cccccc}
\toprule[1.1pt]
 {Method} & {IFEval $\uparrow$} & {GPQA $\uparrow$} & {MMLU $\uparrow$} & {HellaSwag $\uparrow$} & {TruthfulQA $\uparrow$} & {GSM8K $\uparrow$} \\
\midrule[1.1pt]
{IT} & {$73.5$} & {$31.5$} & {$71.8$} & {$62.1$} & {$56.4$} & {$81.7$} \\
\hline
{DPO} & {$73.2$} & {$32.0$} & {$71.9$} & {$62.0$} & {$57.1$} & {$71.3$}  \\
{+\textbf{MaPPO}} & {$\mathbf{73.8}$} & {$\mathbf{33.1}$} & {$\mathbf{72.0}$} & {$\mathbf{62.1}$} & {$\mathbf{59.2}$} 
& {$\mathbf{80.1}$} \\
\hline
{I-DPO} & {$\textbf{72.9}$} & {$33.0$} & {$71.9$} & {$62.2$} & {$55.9$} & {$73.2$} \\
{+\textbf{MaPPO}} & $72.6$ & ${\textbf{33.3}}$ & $\mathbf{72.9}$ & $\mathbf{62.2}$ & $\mathbf{56.2}$ & $\mathbf{82.0}$ \\
\bottomrule[1.1pt]
\end{tabular}
\end{table}

We test the performance on six widely used academic benchmarks, evaluating various model abilities, including explicit instruction following (IFEval) \citep{zhou2023instruction}, general knowledge (GPQA) \citep{rein2024gpqa}, multitask language understanding (MMLU) \citep{hendrycks2021measuring}, commonsense reasoning (HellaSwag) \citep{zellers-etal-2019-hellaswag}, human falsehoods mimicking (TruthfulQA) \citep{lin-etal-2022-truthfulqa}, and math word problem-solving (GSM8K) \citep{cobbe2021training}. 

We show the results on the six academic benchmarks from Llama-3-8B-Instruct model in Table \ref{table:academic}, and Qwen2.5-7B-Instruct model in Table \ref{table:academic_qwen_7b}. In general, for offline DPO with MaPPO outperforms the original DPO in all benchmarks for both models. The improvement is significant on GSM8K for Qwen2.5-7B-Instruct, and on TruthfulQA for Llama-3-8B-Instruct. For the iterative DPO with MaPPO, the performances are better than the original I-DPO on most benchmarks, and maintains the performances on IFEval and GPQA. Overall, the performances of online methods are better than offline methods, and MaPPO generally improves or maintains the  performances on academic benchmarks in both settings.

\subsubsection{Broader Evaluation} 
\label{app:Broader_Evaluation}

\paragraph{Human Evaluation.}
Following the setting in~\citet{ethayarajh2024kto, yang2024using}, we randomly sampled $1000$ prompts from the prompt set and generated outputs from the Qwen2.5-7B-Instruct models optimized with DPO and +MaPPO. Each rater scored the response pair with preferred \& unpreferred label. Each response pair was evaluated by $5$ raters, and a prompt was counted as a win if at least 3 raters preferred the response. 
We then report the winrates generated from Qwen2.5-7B-Instruct with different PO methods. 
Under the $95\%$ confidence interval, the standard error (SE) is $0.0155$, and CI = Mean$\pm$($1.96\times$SE). The results demonstrate the superiority of DPO+MaPPO ($0.594\pm0.0304$) compared to vanilla DPO ($0.406\pm0.0304$).

\paragraph{Evaluation Instructions.}
For each example, evaluators were shown a prompt and two candidate responses, Response A and Response B. Evaluators were instructed to select the response that they preferred based on overall quality, including correctness, relevance, helpfulness, and clarity.

\paragraph{LLM as a Judge Comparison.}
The model performance with a direct head-to-head comparison between the baseline and MaPPO-enhanced models may bring more information than only comparing both against GPT-4. We therefore conducted this additional evaluation with GPT-4.1 as the judge in Table \ref{tab:hth_comparison}. In the DPO vs. DPO+MaPPO comparison, for Qwen2.5-7B-Instruct, MaPPO achieves a $52.86$ win rate, while DPO obtains $47.14$ with the standard error $1.74$.
As expected, these win rates are close to $50$ because the comparison is between two much closer models, but the result still consistently favors MaPPO.

\begin{table}[!ht]
\centering
\caption{Head-to-head comparison of DPO and DPO+MaPPO win rates ($\%$).}
\begin{tabular}{lccc}
\toprule
\textbf{Model} & \textbf{DPO Win Rate} & \textbf{DPO+MaPPO Win Rate} & \textbf{Standard Error} \\ \midrule
Qwen2.5-14B-Instruct    & $46.31$ & $\mathbf{53.69}$ & $1.72$ \\
Qwen2.5-7B-Instruct    & $47.14$ & $\mathbf{52.86}$ & $1.74$ \\
Qwen2.5-3B-Instruct    & $46.53$ & $\mathbf{53.47}$ & $1.52$ \\
Qwen2.5-1.5B-Instruct  & $47.26$ & $\mathbf{52.74}$ & $1.31$ \\ \bottomrule
\end{tabular}
\label{tab:hth_comparison}
\end{table}

\subsubsection{Ablation Study on RMs} 
\label{app:Ablation_RMs}

As illustrated in Equation~\eqref{eq:prior_prob}, the prior signal is instantiated via a reward model (RM).
We formalize a theoretical statement in Appendix \ref{app:rm_equal}: Compared to changing the prior function, changing the reward model could achieve an equivalent effect on the prior signal. Thus, this section provides the empirical results with different RMs.

To better understand the role of RMs, we conduct an ablation study by integrating different open-source RMs into our framework, including BT model\footnote{https://huggingface.co/sfairXC/FsfairX-LLaMA3-RM-v0.1} (sfairXC/FsfairX-LLaMA3-RM-v0.1), MoE model\footnote{https://huggingface.co/RLHFlow/ArmoRM-Llama3-8B-v0.1} (RLHFlow/ArmoRM-Llama3-8B-v0.1), uncertainty-aware (UA) model\footnote{https://huggingface.co/LxzGordon/URM-LLaMa-3.1-8B} (LxzGordon/URM-LLaMa-3.1-8B), and a low-quality (LQ) RM\footnote{https://huggingface.co/weqweasdas/RM-Gemma-2B} (weqweasdas/RM-Gemma-2B). weqweasdas/RM-Gemma-2B is the sixth from the bottom on the RewardBench\footnote{https://huggingface.co/spaces/allenai/reward-bench}. Thus, this RM's quality is not that ideal compared to previous RMs.
We add a standard logistic function on the absolute rewards to make sure $r \in [0,1]$.

As shown in Table~\ref{table:RewardAblation_app}, the absolute performance of I-DPO varies considerably depending on the chosen RM. Importantly, across all settings, our proposed MaPPO plugin consistently boosts performance on AlpacaEval 2.0, Arena-Hard, and MT-Bench. These results highlight two key insights: (i) the choice of RM can influence the strength of supervision and the final outcomes, and (ii) MaPPO is robust to such variation, reliably enhancing DPO regardless of the underlying RM. This robustness underscores the generality and flexibility of our framework.

\begin{table}[!ht]
\caption{Ablation on reward models. $\uparrow$ indicates the higher the better.}
\centering
\label{table:RewardAblation_app}
\resizebox{\textwidth}{!}{%
\begin{tabular}{l|l|l|lll}
\toprule[1.1pt]
{Model} & {Reward Model} & {Method} & {AlpacaEval 2.0 $\uparrow$} & {Arena-Hard $\uparrow$} & {MT-Bench $\uparrow$} \\
\midrule[1.1pt]

\multirow{9}{*}{Qwen2.5-7B-IT}  
& \multirow{3}{*}{BT} 
& {IT} & {$27.03$}  & {$42.9$ }  & {$8.61$ } \\
& & {I-DPO} & {$33.80$} & {$46.9$} & {$\mathbf{8.55}$}\\
& & {+\textbf{MaPPO}} & {$\mathbf{39.10}$ $\textcolor{ForestGreen}{+5.30}$} & {$\mathbf{61.6}$ $\textcolor{ForestGreen}{+14.7}$} & {$8.54$ $\textcolor{Apricot}{-0.01}$}\\
\cline{2-6}
{}
& \multirow{3}{*}{MoE} 
& {IT} & {$27.03$}  & {$42.9$ }  & {$8.61$ } \\
& & {I-DPO} & {$28.79$} & {$48.3$} & {$8.58$}\\
& & {+\textbf{MaPPO}} & {$\mathbf{31.56}$ $\textcolor{ForestGreen}{+2.77}$} & {$\mathbf{52.0}$ $\textcolor{ForestGreen}{+3.7}$} & {$\mathbf{8.60}$ $\textcolor{ForestGreen}{+0.02}$}\\

\cline{2-6}
{}  
& \multirow{3}{*}{UA} 
& {IT} & {$27.03$}  & {$42.9$ }  & {$8.61$ } \\
& & {I-DPO} & {$32.26$} & {$55.1$} & {$8.60$}\\
& & {+\textbf{MaPPO}} & {$\mathbf{36.17}$ $\textcolor{ForestGreen}{+3.91}$} & {$\mathbf{59.4}$ $\textcolor{ForestGreen}{+4.3}$} & {$\mathbf{8.62}$ $\textcolor{ForestGreen}{+0.02}$}\\

\cline{2-6}
{}
& \multirow{3}{*}{low quality} 		
& {IT} & {$27.03$} &{$42.9$}&{$8.61$}\\
& & {I-DPO} & {$27.77$} &{$43.5$ }&{$8.06$}\\
& & {+\textbf{MaPPO}} & {$\mathbf{29.78}$ $\textcolor{ForestGreen}{+2.01}$} & {$\mathbf{48.4}$ $\textcolor{ForestGreen}{+4.9}$} & {$\mathbf{8.63}$ $\textcolor{ForestGreen}{+0.57}$} \\

\midrule[1.1pt]

\multirow{9}{*}{Llama-3-8B-IT}  
& \multirow{3}{*}{BT} 
& {IT} & {$10.85$} &{$10.2$}&{$7.52$}\\
& & {I-DPO} & {${29.47}$} &{$25.6$ }&{$8.01$}\\
& & {+\textbf{MaPPO}} & {$\mathbf{32.68}$ $\textcolor{ForestGreen}{+3.21}$} & {$\mathbf{31.0}$ $\textcolor{ForestGreen}{+5.4}$} & {$\mathbf{8.04}$ $\textcolor{ForestGreen}{+0.03}$} \\
\cline{2-6}
{}  
& \multirow{3}{*}{MoE} 
& {IT} & {$10.85$} &{$10.2$}&{$7.52$}\\
& & {I-DPO} & {$12.69$} &{$8.1$ }&{$7.44$}\\
& & {+\textbf{MaPPO}} & {$\mathbf{16.69}$ $\textcolor{ForestGreen}{+4.00}$} & {$\mathbf{16.4}$ $\textcolor{ForestGreen}{+8.3}$} & {$\mathbf{7.78}$ $\textcolor{ForestGreen}{+0.34}$} \\

\cline{2-6}
{}  
& \multirow{3}{*}{UA} 
& {IT} & {$10.85$} &{$10.2$}&{$7.52$}\\
& & {I-DPO} & {$25.61$} &{$30.4$ }&{$8.06$}\\
& & {+\textbf{MaPPO}} & {$\mathbf{27.83}$ $\textcolor{ForestGreen}{+2.22}$} & {$\mathbf{33.7}$ $\textcolor{ForestGreen}{+3.3}$} & {$\mathbf{8.10}$ $\textcolor{ForestGreen}{+0.04}$} \\

\cline{2-6}
{}  
& \multirow{3}{*}{low quality} 
& {IT} & {$10.85$} &{$10.2$}&{$7.52$}\\
& & {I-DPO} & {$11.24$} &{$13.4$ }&{$7.96$}\\
& & {+\textbf{MaPPO}} & {$\mathbf{15.76}$ $\textcolor{ForestGreen}{+4.52}$} & {$\mathbf{16.1}$ $\textcolor{ForestGreen}{+2.7}$} & {$\mathbf{8.01}$ $\textcolor{ForestGreen}{+0.05}$} \\

\bottomrule[1.1pt]
\end{tabular}
} 
\end{table}

\begin{table}[!ht]
\caption{Comparison w/wo SFT-B and w/wo MaPPO (+MaP). $\uparrow$ indicates the higher the better.}
\centering
  \small
\label{table:EvaluationICLR_app}
\begin{tabular}{l|l|lll}
\toprule[1.1pt]
{Model} & {Method} & {AlpacaEval 2.0 $\uparrow$} & {Arena-Hard $\uparrow$} & {MT-Bench $\uparrow$} \\
\midrule[1.1pt]
\multirow{5}{*}{Llama-3-8B-IT} & {IT} & {$10.85$} &{$10.2$}&{$7.52$}\\
\cline{2-5}
{ } & {I-DPO} & {${29.47}$} &{$25.6$}&{$8.01$} \\
{ } & {+\textbf{MaP}} & {$\mathbf{32.68}$ $\textcolor{ForestGreen}{+3.21}$ $\textcolor{blue}{+3.21}$} & {$\mathbf{31.0}$ $\textcolor{ForestGreen}{+5.4}$ $\textcolor{blue}{+5.4}$} & {$\mathbf{8.04}$ $\textcolor{ForestGreen}{+0.03}$ $\textcolor{blue}{+0.03}$} \\
\cline{2-5}
{ } & {+\textbf{SFT-B}} & {$35.55$ ~$\textcolor{ForestGreen}{+6.08}$} & {$30.1$ ~$\textcolor{ForestGreen}{+4.5}$} & {$8.02$ ~$\textcolor{ForestGreen}{+0.01}$} \\
{ } & {+\textbf{SFT-B +MaP}} & {$\mathbf{36.21}$ $\textcolor{ForestGreen}{+6.74}$ $\textcolor{blue}{+0.66}$} & {$\mathbf{31.5}$ $\textcolor{ForestGreen}{+5.9}$ $\textcolor{blue}{+1.4}$} & {$\mathbf{8.03}$ $\textcolor{ForestGreen}{+0.02}$ $\textcolor{blue}{+0.01}$} \\
\midrule[1.1pt]
\multirow{5}{*}{Qwen2.5-7B-IT} & {IT} & {$27.03$}  & {$42.9$ }  & {$8.61$ } \\
\cline{2-5}
{ } & {I-DPO} & {$33.80$} & {$46.9$} & {$\mathbf{8.55}$}\\
{ } & {+\textbf{MaP}} & {$\mathbf{39.10}$ $\textcolor{ForestGreen}{+5.30}$ ~$\textcolor{blue}{+5.30}$} & {$\mathbf{61.6}$ $\textcolor{ForestGreen}{+14.7}$ $\textcolor{blue}{+14.7}$} & {$8.54$ $\textcolor{Apricot}{-0.01}$ $\textcolor{blue}{-0.01}$} \\
\cline{2-5}
{ } & {+\textbf{SFT-B}} & {$51.36$\ \ $\textcolor{ForestGreen}{+17.56}$} & {$33.5$\ \ $\textcolor{Apricot}{-13.4}$} & {$8.01$ $\textcolor{Apricot}{-0.54}$} \\
{ } & {+\textbf{SFT-B +MaP}} & {$\textbf{55.84}$ $\textcolor{ForestGreen}{+22.04}$ $\textcolor{blue}{+4.48}$} & {$\mathbf{43.5}$ ~$\textcolor{Apricot}{-3.4}$ ~$\textcolor{blue}{+10.0}$} & {$\mathbf{8.40}$ $\textcolor{Apricot}{-0.15}$ $\textcolor{blue}{+0.39}$}\\
\midrule[1.1pt]
\multirow{5}{*}{Qwen2.5-14B-IT} & {IT} & {$31.09$}  & {$43.6$ }  & {$8.62$ } \\
\cline{2-5}
{ } & {I-DPO} & {$45.07$} & {$54.5$} & {${8.64}$}\\
{ } & {+\textbf{MaP}} & {$\mathbf{48.89}$ $\textcolor{ForestGreen}{+3.82}$ ~$\textcolor{blue}{+3.82}$} & {$\mathbf{62.1}$ $\textcolor{ForestGreen}{+7.6}$ $\textcolor{blue}{+7.6}$} & {$\mathbf{8.80}$ $\textcolor{ForestGreen}{+0.16}$ $\textcolor{blue}{+0.16}$} \\
\cline{2-5}
{ } & {+\textbf{SFT-B}} & {$54.60$\ \ $\textcolor{ForestGreen}{+9.53}$} & {$40.3$\ \ $\textcolor{Apricot}{-14.2}$} & {$8.51$ $\textcolor{Apricot}{-0.13}$} \\
{ } & {+\textbf{SFT-B +MaP}} & {$\textbf{57.13}$ $\textcolor{ForestGreen}{+12.06}$ $\textcolor{blue}{+2.53}$} & {$\mathbf{45.5}$ ~$\textcolor{Apricot}{-9.0}$ ~$\textcolor{blue}{+5.2}$} & {$\mathbf{8.52}$ $\textcolor{Apricot}{-0.12}$ $\textcolor{blue}{+0.01}$}\\
\bottomrule[1.1pt]
\end{tabular}
\end{table}

\subsubsection{Comparison and Compatibility with Other Methods} 
\label{app:Comparison}

We also compare our approach with complementary techniques, such as the method proposed by \cite{ren2024learning}, which introduces an additional SFT-based pre-processing (denoted as SFT-B) to mitigate the squeezing effect.
It is worthy to mention that our proposed method is \textbf{compatible} with SFT-B.

In Table~\ref{table:EvaluationICLR_app}, we use the \textcolor{blue}{blue} color to indicate the improvement with the MaPPO plugin, and the \textcolor{ForestGreen}{green} color to indicate the improvement compared to I-DPO. 
We find SFT-B alone can bring substantial gains in certain metrics but may also introduce trade-offs on others (e.g., lower Arena-Hard performance on Qwen2.5-7B-Instruct). By contrast, when combined with MaPPO, the two methods operate jointly and achieve consistent improvements across most benchmarks. These findings demonstrate that MaPPO can be seamlessly integrated with diverse training strategies, further broadening its applicability in practice.

\subsubsection{Direct Validation of the Near-Tie Calibration Mechanism}
\label{app:near_tie_validation}

To directly examine whether MaPPO mitigates the squeezing effect for near-tie preference pairs, we conduct a reward-gap-stratified analysis on the same held-out preference pairs for DPO and MaPPO. For each response $y$, we compute its token-normalized log-likelihood under policy $\pi$ as
\begin{equation}
\bar{\ell}_{\pi}(y\mid x)
=
\frac{1}{|y|}
\sum_{t=1}^{|y|}
\log \pi(y_t\mid x,y_{<t}).
\end{equation}
Token normalization prevents response length from dominating the comparison. We partition the held-out pairs according to the external reward gap
$\Delta_r=r_w-r_l$
using four bins:
$[0,0.1]$, $(0.1,0.3]$, $(0.3,0.5]$, and $(0.5,1]$.

\paragraph{Rejected-response gradient.}
We first evaluate DPO and MaPPO at the same initial policy checkpoint using identical preference pairs. For each pair, the magnitudes of the rejected-response gradient components are
\begin{equation}
G_l^{\mathrm{DPO}}
=
\beta(1-\sigma(u_{\mathrm{DPO}}))
\left\|
\nabla_\theta \log \pi_\theta(y_l\mid x)
\right\|,
\end{equation}
and
\begin{equation}
G_l^{\mathrm{MaPPO}}
=
\beta\Delta_r(1-\sigma(u_{\mathrm{MaPPO}}))
\left\|
\nabla_\theta \log \pi_\theta(y_l\mid x)
\right\|.
\end{equation}
This diagnostic captures both the explicit coefficient $\Delta_r$ and its effect through the logistic factor, and therefore measures the actual rejected-response gradient component.

\begin{table}[t]
\centering
\caption{Post-training likelihood dynamics across reward-gap bins.}
\label{tab:near_tie_likelihood}
\begin{tabular}{llrrrr}
\toprule
Reward-gap bin & Method &
$\mathbb{E}[\delta_w]$ &
$\mathbb{E}[\delta_l]$ &
Chosen-drop rate &
Squeezing rate \\
\midrule
$[0,0.1]$
& DPO   & $-0.183$ & $-0.435$ & 78.2\% & 75.9\% \\
& MaPPO & $+0.030$ & $-0.076$ & 23.0\% & 22.6\% \\
\midrule
$(0.1,0.3]$
& DPO   & $-0.156$ & $-0.503$ & 57.0\% & 55.4\% \\
& MaPPO & $-0.008$ & $-0.249$ & 15.6\% & 15.5\% \\
\midrule
$(0.3,0.5]$
& DPO   & $-0.126$ & $-0.577$ & 11.3\% & 11.0\% \\
& MaPPO & $-0.050$ & $-0.434$ & 8.4\% & 8.2\% \\
\midrule
$(0.5,1]$
& DPO   & $-0.091$ & $-0.687$ & 6.5\% & 6.5\% \\
& MaPPO & $-0.085$ & $-0.648$ & 4.5\% & 4.4\% \\
\bottomrule
\end{tabular}
\end{table}

The attenuation is strongest for near-tie pairs: MaPPO reduces the rejected-response gradient by $94.4\%$ in the smallest-gap bin, while the reduction progressively decreases to $23.5\%$ in the largest-gap bin. This behavior directly confirms that MaPPO applies stronger calibration when the two responses have similar rewards and increasingly approaches the DPO update as the reward gap grows.

\paragraph{Post-training likelihood dynamics.}
We further compare the trained policies against the initial instruction-tuned policy $\pi_{\mathrm{IT}}$. For $M\in\{\mathrm{DPO},\mathrm{MaPPO}\}$, define
\begin{equation}
\delta_w^M
=
\bar{\ell}_{\pi_M}(y_w\mid x)
-
\bar{\ell}_{\pi_{\mathrm{IT}}}(y_w\mid x),
\qquad
\delta_l^M
=
\bar{\ell}_{\pi_M}(y_l\mid x)
-
\bar{\ell}_{\pi_{\mathrm{IT}}}(y_l\mid x),
\end{equation}
and define the squeezing rate as
\begin{equation}
S^M
=
\Pr\left(
\delta_w^M<0
\wedge
\delta_l^M<0
\right).
\end{equation}

The largest difference between DPO and MaPPO occurs for near-tie pairs. In the $[0,0.1]$ reward-gap bin, MaPPO changes the average preferred-response log-likelihood from $-0.183$ to $+0.030$ and reduces the squeezing rate from $75.9\%$ to $22.6\%$. As the reward gap increases, the difference between DPO and MaPPO becomes progressively smaller, consistent with MaPPO increasingly approaching DPO for clearly separated pairs.

\subsubsection{Comparison with Reward-Aware and RL-Style Baselines}
\label{app:reward_aware_baselines}

We further compare MaPPO against reward-aware and RL-style alternatives. 
We consider Nemotron-4 RPO as a reward-margin baseline, RLOO as an RL-style scalar-reward baseline, and PPO and GRPO as policy-gradient-based RL baselines.

\paragraph{Reward-aware and online baselines.}
For controlled comparisons on Qwen2.5-7B-Instruct, RPO, RLOO, DPO, and MaPPO use the same initial policy, prompt set, reward model, generation budget, training budget, and evaluation pipeline whenever applicable.

We note that RPO is not fundamentally restricted to offline training. When new responses and reward-model scores are available at each iteration, its target reward gap can be recomputed, and RPO can in principle be incorporated into an iterative preference-optimization pipeline. The main distinction is therefore in how quantitative reward information enters the objective: RPO encourages the policy's implicit reward gap to match an externally specified reward margin, whereas MaPPO directly uses the external reward gap to calibrate the rejected-response contribution.

\begin{table}[t]
\centering
\caption{Comparison with reward-aware and RL-style baselines on Qwen2.5-7B-Instruct.}
\label{tab:reward_aware_baselines}
\begin{tabular}{lccc}
\toprule
Method & AlpacaEval 2.0 LC WR & Arena-Hard & MT-Bench \\
\midrule
IT                  & 27.03 & 42.9 & 8.61 \\
DPO                 & 32.01 & 45.5 & 8.56 \\
RPO                 & 32.76 & 46.2 & 8.56 \\
DPO + MaPPO         & 38.24 & 59.2 & 8.79 \\
RLOO                & 33.12 & 46.7 & 8.64 \\
I-DPO               & 33.80 & 46.9 & 8.55 \\
I-RPO               & 34.82 & 56.7 & 8.54 \\
I-DPO + MaPPO       & 39.10 & 61.6 & 8.54 \\
\bottomrule
\end{tabular}
\end{table}

Under the matched offline setting, RPO improves over DPO by 0.75 points on AlpacaEval 2.0 and 0.7 points on Arena-Hard, whereas DPO + MaPPO improves over DPO by 6.23 and 13.7 points, respectively. Compared directly with RPO, DPO + MaPPO improves AlpacaEval 2.0 by 5.48 points and Arena-Hard by 13.0 points.

The same trend holds in the iterative setting. I-DPO + MaPPO improves over I-RPO by 4.28 points on AlpacaEval 2.0 and 4.9 points on Arena-Hard, while achieving the same MT-Bench score. Compared with RLOO, I-DPO + MaPPO improves AlpacaEval 2.0 by 5.98 points and Arena-Hard by 14.9 points.

\paragraph{Policy-gradient RL baselines.}
We additionally compare against PPO and GRPO on Qwen2.5-14B-Instruct. These methods belong to a different optimization paradigm and require online rollouts and reward-driven policy-gradient updates, but provide complementary RL-style baselines. We use a clip ratio $\epsilon=0.2$ for GRPO and otherwise follow the same evaluation protocol. For AlpacaEval 2.0, the average standard error is 1.15; for Arena-Hard, the confidence interval is approximately $\pm 1.5$ points; and the MT-Bench score is averaged over three judge runs.

\begin{table}[t]
\centering
\caption{Performance comparison with policy-gradient RL baselines on Qwen2.5-14B-Instruct.}
\label{tab:policy_gradient_baselines}
\begin{tabular}{lccc}
\toprule
Method & AlpacaEval 2.0 $\uparrow$ & Arena-Hard $\uparrow$ & MT-Bench $\uparrow$ \\
\midrule
GRPO                & 47.17 & 56.9 & 8.65 \\
PPO                 & 48.13 & 57.5 & 8.61 \\
I-DPO + MaPPO       & 48.89 & 62.1 & 8.80 \\
\bottomrule
\end{tabular}
\end{table}

I-DPO + MaPPO achieves higher performance than both PPO and GRPO on AlpacaEval 2.0 and Arena-Hard while maintaining comparable MT-Bench performance. Together with the RLOO and RPO comparisons above, these results show that MaPPO remains effective against iterative preference optimization, reward-margin-based optimization, and policy-gradient-based RL baselines. They also suggest that access to quantitative reward information alone does not explain the observed gains; how this information is incorporated into the optimization objective is important.

\subsubsection{Length Analysis and Practical Costs}
\label{app:length_cost}

We additionally report the raw AlpacaEval 2.0 win rate, length-controlled win rate, and average response length for the Qwen2.5-7B-Instruct iterative setting.

\begin{table}[t]
\centering
\caption{Raw and length-controlled AlpacaEval 2.0 results.}
\label{tab:alpaca_length}
\begin{tabular}{lccc}
\toprule
Method & Raw WR & LC WR & Avg.\ response length \\
\midrule
I-DPO          & 57.04 & 33.80 & $\sim$3.2k tokens \\
I-DPO + MaPPO  & 62.10 & 39.10 & $\sim$3.1k tokens \\
\bottomrule
\end{tabular}
\end{table}

MaPPO improves the raw win rate from 57.04 to 62.10, an absolute gain of 5.06 points. The improvement remains after controlling for response length, with the length-controlled win rate increasing from 33.80 to 39.10, an absolute gain of 5.30 points. Meanwhile, the average response length slightly decreases from approximately 3.2k to 3.1k tokens. Thus, the improvement is not attributable to generating longer responses.

\paragraph{Computational and reward-model costs.}
MaPPO uses the same policy forward and backward passes as I-DPO. Its additional policy-training computation only requires computing the scalar reward gap and multiplying the rejected-response log-ratio by this scalar. In our iterative pipeline, the reward model is already used to score and rank candidate responses, and MaPPO directly reuses these scores without additional reward-model calls.

\begin{table}[t]
\centering
\caption{Average practical cost per global batch under the same training configuration with 8 NVIDIA H100 GPUs.}
\label{tab:practical_cost}
\begin{tabular}{lccc}
\toprule
Method & Training (s) & RM scoring (s) & Peak memory \\
\midrule
I-DPO          & 14 & 39 & 41\% \\
I-DPO + MaPPO  & 14 & 39 & 41\% \\
RLOO           & 29 & 39 & 69\% \\
PPO            & 55 & 39 & 92\% \\
\bottomrule
\end{tabular}
\end{table}

Under this configuration, MaPPO introduces no measurable additional policy-training time, reward-model scoring time, or training-engine peak memory relative to I-DPO. Because the reward scores required by MaPPO are already computed for response ranking in our iterative pipeline, its incremental reward-model inference cost is zero. For an offline human-labeled preference dataset without scalar reward scores, MaPPO instead requires a one-time reward-model scoring pass over the chosen and rejected responses.
\section{Further Discussions}
\label{app:Discussions}

\subsection{Prior Signal}
\label{app:Prior_Function}

In Bayes estimation, the prior distribution is usually constructed by experts with domain knowledge without an exception to avoid it \citep{jaynes2007prior}.
We choose a simple form in Equation~\eqref{eq:prior_prob}, which has the same structure as the widely used prior function \citep{blei2010probabilistic} and aligns with the softmax probability. 
This brings a very clean result in Equation~\eqref{eq:mappo_loss}.
In this paper, our goal is to introduce the MaP-based preference optimization framework and present a simple, effective prior that is compatible with the BT/softmax formulation, rather than to explore the space of possible prior designs.
Other function choices that are designed by domain experts are also acceptable and open to be used, but will lose the clean form mathematically and thus make then hard to compute in the loss function.

Thus, instead of changing the prior function, we argue that changing the reward model could achieve an equivalent effect on the prior signal.

Specifically, in our current formulation, the prior affects MaPPO through a scalar calibration coefficient on the rejected-response term. Therefore, for a broad class of alternative prior designs that only modify this scalar coefficient, changing the prior design is exactly \textbf{equivalent} to changing the effective calibration signal used by MaPPO. 
Moreover, when the alternative design is difference-separable, i.e., it can be written as $g(r_{w}, r_{l}) = T(r_{w}) - T(r_{l})$, it is exactly equivalent to keeping the same MaPPO prior form and replacing the reward model by a transformed reward model $\tilde r = T \circ r$. We formalize this theoretical statement in Appendix \ref{app:rm_equal}.
Hence, an ablation over multiple reward models would cover a meaningful family of prior-function variations. 

Empirically, we perform ablation studies on reward models in Appendix \ref{app:supp_results}. Table \ref{table:RewardAblation_app} (with low-quality RM results) shows that although the absolute performance varies with the reward model, MaPPO consistently improves over the corresponding baseline across settings, suggesting that the proposed MaP principle is robust to different prior signals.

\subsection{Relation to Recent Policy-Gradient-Based RL Methods}
\label{app:recent_rl_methods}

In addition to DPO-style preference optimization, recent policy-gradient-based methods explore complementary directions for LLM post-training. DAPO~\cite{yu2025dapo} introduces a scalable RL framework incorporating data selection, asymmetric clipping, overlong reward shaping, and token-level policy-gradient losses. GSPO~\cite{zheng2025group} extends sequence-level policy optimization to groupwise signals and large-scale mixture-of-experts training, while PRO~\cite{guo2025proximalized} develops a proximalized formulation that decomposes preference signals into stable update components.

These approaches primarily improve the policy-gradient optimization pipeline, whereas MaPPO modifies the DPO-style preference objective by incorporating reward-gap information into the rejected-response contribution. The two directions are therefore complementary, and incorporating MaPPO-style reward calibration into policy-gradient-based methods is an interesting direction for future work. Quantitative comparisons with PPO, GRPO, and RLOO are provided in Appendix~\ref{app:reward_aware_baselines}.

\clearpage
\newpage
\section{Supplementary Examples}
\label{app:Supplementary_Examples}

\begin{figure}[!htbp]
\begin{tcolorbox}[title=Prompt~$\mathbf{x}$,fonttitle=\bfseries, boxrule=0.5mm]
Solve the following math problem step-by-step. Simplify your answer as much as possible. Present your final answer as \boxed{Your Answer}. If the sum of two positive integers is 22 and the difference of their squares is 44, what is the product of the two integers?
\end{tcolorbox}
\end{figure}

\subsection{Before Preference Optimization}
\begin{figure}[!htbp]
\begin{tcolorbox}[colback=green!5, colframe=black!60!green!45, title={\textbf{Response}~$\mathbf{y}_{w}$, $r=0.93$}, fonttitle=\bfseries, boxrule=0.5mm]
To solve this problem step-by-step, let's denote the two positive integers as \(x\) and \(y\) with \(x > y\).

Step 1: Set Up Equations

~~~~Given:

~~~~~~~~1. The sum of the two integers: \(x + y = 22\)

~~~~~~~~2. The difference of their squares: \(x^2 - y^2 = 44\)
\[ \]

Step 2: Factorize the Difference of Squares
Recall that the difference of squares can be factored as:
\[ x^2 - y^2 = (x + y)(x - y) \]
Substitute the given sum \(x + y = 22\) into the factored form:
\[ 44 = 22(x - y) \]

Step 3: Solve for \(x - y\)
Divide both sides by 22:
\[ x - y = 2 \]

Step 4: Solve the System of Linear Equations
Now, we have the system:
1. \(x + y = 22\)
2. \(x - y = 2\)

Add these two equations to eliminate \(y\):
\[ (x + y) + (x - y) = 22 + 2 \]
\[ 2x = 24 \]
\[ x = 12 \]

Substitute \(x = 12\) back into the first equation to find \(y\):
\[ 12 + y = 22 \]
\[ y = 10 \]

Step 5: Calculate the Product
The product of the two integers \(x\) and \(y\) is:
\[ x \cdot y = 12 \cdot 10 = 120 \]

Final Answer
\[ \boxed{120} \] 
\end{tcolorbox}
\end{figure}

\begin{figure}[!htbp]
\begin{tcolorbox}[colback=red!5, colframe=black!60!red!45, title={\textbf{Response}~$\mathbf{y}_{l}$, $r=0.90$}, fonttitle=\bfseries, boxrule=0.5mm]
To solve this problem step-by-step, let's denote the two positive integers as \(x\) and \(y\), where \(x > y\). We are given two conditions:

~~~~1. The sum of the two integers is 22:
   \[ x + y = 22 \]

~~~~2. The difference of their squares is 44:
   \[ x^2 - y^2 = 44 \]

We can use the identity for the difference of squares, \(x^2 - y^2 = (x + y)(x - y)\), to rewrite the second equation. Substituting \(x + y = 22\) into the identity gives:
\[
(x + y)(x - y) = 44 \implies 22(x - y) = 44
\]
Dividing both sides by 22:
\[
x - y = 2
\]

Now we have a system of linear equations:
\[
\begin{cases}
x + y = 22 \\
x - y = 2
\end{cases}
\]

Adding these two equations eliminates \(y\):
\[
(x + y) + (x - y) = 22 + 2 \implies 2x = 24 \implies x = 12
\]

Substituting \(x = 12\) back into the first equation to solve for \(y\):
\[
12 + y = 22 \implies y = 10
\]

Thus, the two integers are \(x = 12\) and \(y = 10\).

To find the product of the two integers:
\[ xy = 12 \times 10 = 120 \]

Therefore, the product of the two integers is \boxed{120}. 
\end{tcolorbox}
\begin{center}
\caption{Before MLE optimization, the model consistently generates high-quality (high rewards) answers $\mathbf{y}_{w}$ and $\mathbf{y}_{l}$ with prompt $\mathbf{x}$.}
\end{center}
\end{figure}

\clearpage
\newpage
\subsection{After MLE Optimization}

\begin{figure}[!htbp]
\begin{tcolorbox}[breakable,colback=green!5, colframe=black!60!green!45, title={\textbf{Response}~$\mathbf{y}_{w}$, $r=0.87$}, fonttitle=\bfseries, boxrule=0.5mm]
To solve the given problem, we'll proceed through the following steps systematically:

Given:

~~~~1. The sum of two positive integers \( x \) and \( y \) is 22.

~~~~2. The difference of their squares is 44.

We need to determine the product of these two integers.

Step-by-Step Solution:

1. Setting up Equations:
   
~~~~- From the first condition: 
     \[
     x + y = 22
     \]
~~~~- From the second condition (expressing the difference of squares):
     \[
     x^2 - y^2 = 44
     \]
   
   We know that \( x^2 - y^2 \) can be factored using the difference of squares formula:
     \[
     x^2 - y^2 = (x+y)(x-y)
     \]

2. Substitute \( x + y \):
   
~~~~- Using the known value from equation (1) in the factored form of the difference of squares:
     \[
     x^2 - y^2 = (x+y)(x-y) \implies 44 = 22(x-y)
     \]
~~~~- Solve for \( x - y \):
     \[
     22(x - y) = 44 \implies x - y = 2
     \]

3. Solving the System of Linear Equations:
   
~~~~- Now, we have the system of equations:
     \[
     x + y = 22
     \]
     \[
     x - y = 2
     \]
~~~~- Add these two equations to eliminate \( y \):
     \[
     (x + y) + (x - y) = 22 + 2
     \]
     \[
     2x = 24 \implies x = 12
     \]
~~~~- Substitute \( x = 12 \) back into \( x + y = 22 \) to find \( y \):
     \[
     12 + y = 22 \implies y = 10
     \]

4. Finding the Product:
   
~~~~- With \( x = 12 \) and \( y = 10 \), calculate the product:
     \[
     xy = 12 \times 10 = 120
     \]

Thus, the product of the two integers is \boxed{120}.
\end{tcolorbox}
\end{figure}

\begin{figure}[!htbp]
\begin{tcolorbox}[colback=red!5, colframe=black!60!red!45, title={\textbf{Response}~$\mathbf{y}_{l}$, $r=0.43$}, fonttitle=\bfseries, boxrule=0.5mm]
Step 1: Interpret the problem and set up equations based on the given information.
Given that the sum of two positive integers is 22 and the difference of their squares is 44, we need to find these integers and subsequently their product.

Let's denote the two integers as \(a\) and \(b\) such that \(a > b\), to simplify handling the difference and sum relationships given directly without specifying which integer represents \(a\) or \(b\) in terms of being greater than the other directly addresses addressing conditions related directly to interpreting the context regarding how the operations (sum, difference of squares) relate between two specified integers, based on the given descriptions for solving in the context presented, focusing on correctly interpreting and directly addressing the specific requirements as outlined for accurately proceeding with solving based on the given details, to correctly proceed according to provided instructions and the nature of solving based on the specifications given for dealing with the described conditions around the operation outcomes involving the integers in question.
\[ \]
Step 2: Translate the given conditions into mathematical equations.

~~~~- The sum of the two integers equals 22: \(a + b = 22\).

~~~~- The difference of their squares equals 44: \(a^2 - b^2 = 44\).
\[ \]
Step 3: Utilize algebraic identities for simplification.
From the second equation, apply the algebraic identity for the difference of squares: 
\[a^2 - b^2 = (a + b)(a - b)\]
Thus, substituting in the known value for \(a + b\):
\[(a + b)(a - b) = 44 \implies 22(a - b) = 44.\]

Step 4: Solve for \(a - b\).
Divide both sides by 22:
\[a - b = 2.\]

Step 5: Solve the system of linear equations.
We now have two simultaneous equations:

~~~~1. \(a + b = 22\)

~~~~2. \(a - b = 2\)

Add the two equations together to eliminate \(b\):
\[(a + b) + (a - b) = 22 + 2 \implies 2a = 24 \implies a = 12.\]

Substitute \(a = 12\) back into the first equation to solve for \(b\):
\[12 + b = 22 \implies b = 10.\]

Step 6: Find the product of \(a\) and \(b\).
The product of the two integers \(a\) and \(b\) is:
\[ab = 12 \times 10 = 120.\]

Step 7: Box the final answer.
\[\boxed{120}\]

\end{tcolorbox}
\begin{center}
\caption{After MLE optimization, the model degenerates, and the outputs $\mathbf{y}_{w}$ and $\mathbf{y}_{l}$ become verbose (low rewards) with prompt $\mathbf{x}$.}
\end{center}
\end{figure}

\clearpage
\newpage
\subsection{After MaP Optimization}

\begin{figure}[!htbp]
\begin{tcolorbox}[colback=green!5, colframe=black!60!green!45, title={\textbf{Response}~$\mathbf{y}_{w}$, $r=0.94$}, fonttitle=\bfseries, boxrule=0.5mm]
To solve this problem step-by-step, let's denote the two positive integers by \(x\) and \(y\), where without loss of generality, we assume \(x > y\).

Given:

~~~~1. The sum of the two integers: \(x + y = 22\)

~~~~2. The difference of their squares: \(x^2 - y^2 = 44\)
\[\]
Step 1: Express the Difference of Squares

~~~~We know from algebra that:
\[ x^2 - y^2 = (x + y)(x - y) \]

~~~~Given \(x + y = 22\), we substitute this into the equation:
\[ 44 = 22(x - y) \]

Step 2: Solve for \(x - y\)

~~~~Divide both sides by 22:
\[ x - y = 2 \]

Step 3: Solve the System of Linear Equations

~~~~Now, we have a system of linear equations:

~~~~~~~~1. \(x + y = 22\)

~~~~~~~~2. \(x - y = 2\)

~~~~Add these two equations to eliminate \(y\):
\[ (x + y) + (x - y) = 22 + 2 \]
\[ 2x = 24 \]
\[ x = 12 \]

~~~~Subtract the second equation from the first to eliminate \(x\):
\[ (x + y) - (x - y) = 22 - 2 \]
\[ 2y = 20 \]
\[ y = 10 \]

Step 4: Calculate the Product of the Two Integers

~~~~Now that we have \(x = 12\) and \(y = 10\), we find their product:
\[ xy = 12 \times 10 = 120 \]

~~~~Thus, the product of the two integers is:
\[\boxed{120}\]
\end{tcolorbox}
\end{figure}

\begin{figure}[!htbp]
\begin{tcolorbox}[colback=red!5, colframe=black!60!red!45, title={\textbf{Response}~$\mathbf{y}_{l}$, $r=0.88$}, fonttitle=\bfseries, boxrule=0.5mm]
To solve this problem, we'll use algebraic methods to represent the given conditions and then solve for the integers. Here are the steps:

1. Define Variables:
   
~~~~Let the two positive integers be \(x\) and \(y\), with \(x > y\).
\[\]
2. Set Up Equations Based on Given Information:
   
~~~~- The sum of the two integers is 22:
     \[
     x + y = 22
     \]
   
~~~~- The difference of their squares is 44:
     \[
     x^2 - y^2 = 44
     \]

3. Factor the Difference of Squares:
   
~~~~Recall that \(x^2 - y^2\) can be factored as \((x + y)(x - y)\). Therefore:
   \[
   (x + y)(x - y) = 44
   \]
~~~~Substitute \(x + y = 22\) into the equation:
   \[
   22(x - y) = 44
   \]

4. Solve for \(x - y\):
   
~~~~Divide both sides by 22:
   \[
   x - y = 2
   \]

5. Set Up a System of Linear Equations:

~~~~We now have the system:
   \[
   \begin{cases}
   x + y = 22 \\
   x - y = 2
   \end{cases}
   \]

6. Add the Two Equations to Solve for \(x\):

~~~~Adding the equations eliminates \(y\):
   \[
   (x + y) + (x - y) = 22 + 2
   \]
~~~~Simplifies to:
   \[
   2x = 24 \implies x = 12
   \]

7. Substitute \(x = 12\) Back to Find \(y\):
   
~~~~Using \(x + y = 22\):
   \[
   12 + y = 22 \implies y = 10
   \]

8. Calculate the Product of the Two Integers:

~~~~The product of \(x\) and \(y\) is:
   \[
   x \cdot y = 12 \cdot 10 = 120
   \]
The answer is
 \[\boxed{120}\]

\end{tcolorbox}
\begin{center}
\caption{After MaP optimization, the model consistently generates high quality outputs with prompt $\mathbf{x}$.}
\end{center}
\end{figure}

\newpage
\section*{NeurIPS Paper Checklist}



\begin{enumerate}

\item {\bf Claims}
    \item[] Question: Do the main claims made in the abstract and introduction accurately reflect the paper's contributions and scope?
    \item[] Answer: \answerYes{} 
    \item[] Justification: The main claims made in the abstract and introduction accurately reflect the paper's contributions and scope.

\item {\bf Limitations}
    \item[] Question: Does the paper discuss the limitations of the work performed by the authors?
    \item[] Answer: \answerYes{} 
    \item[] Justification: The paper discusses the limitations of the work performed by the authors in Section \ref{Sec:Discussions}.

\item {\bf Theory assumptions and proofs}
    \item[] Question: For each theoretical result, does the paper provide the full set of assumptions and a complete (and correct) proof?
    \item[] Answer: \answerYes{} 
    \item[] Justification: For each theoretical result, the paper provides the full set of assumptions and a complete (and correct) proof in Appendix \ref{app:Theoretical_Results}.

    \item {\bf Experimental result reproducibility}
    \item[] Question: Does the paper fully disclose all the information needed to reproduce the main experimental results of the paper to the extent that it affects the main claims and/or conclusions of the paper (regardless of whether the code and data are provided or not)?
    \item[] Answer: \answerYes{} 
    \item[] Justification: The paper fully discloses all the information needed to reproduce the main experimental results of the paper in \ref{sec:Experiments_setup} and Appendix \ref{sec:app_exp_setting}.

\item {\bf Open access to data and code}
    \item[] Question: Does the paper provide open access to the data and code, with sufficient instructions to faithfully reproduce the main experimental results, as described in supplemental material?
    \item[] Answer: \answerYes{} 
    \item[] Justification: The data is publicly available and specified in Section \ref{sec:Experiments_setup}. The complete code will be released before publication.

\item {\bf Experimental setting/details}
    \item[] Question: Does the paper specify all the training and test details (e.g., data splits, hyperparameters, how they were chosen, type of optimizer) necessary to understand the results?
    \item[] Answer: \answerYes{} 
    \item[] Justification: The paper specifies all the training and test details in Section \ref{sec:Experiments_setup} and Appendix \ref{sec:app_exp_setting}.

\item {\bf Experiment statistical significance}
    \item[] Question: Does the paper report error bars suitably and correctly defined or other appropriate information about the statistical significance of the experiments?
    \item[] Answer: \answerYes{} 
    \item[] Justification: The paper reports standard errors and confidence intervals in Section \ref{sec:Experiments:Main_Results}, and Appendices \ref{app:Broader_Evaluation} and \ref{app:reward_aware_baselines}.

\item {\bf Experiments compute resources}
    \item[] Question: For each experiment, does the paper provide sufficient information on the computer resources (type of compute workers, memory, time of execution) needed to reproduce the experiments?
    \item[] Answer: \answerYes{} 
    \item[] Justification: The paper provides sufficient information on the computer resources in Appendix \ref{sec:app_exp_setting}.
    
\item {\bf Code of ethics}
    \item[] Question: Does the research conducted in the paper conform, in every respect, with the NeurIPS Code of Ethics \url{https://neurips.cc/public/EthicsGuidelines}?
    \item[] Answer: \answerYes{} 
    \item[] Justification: The research conducted in the paper conforms, in every respect, with the NeurIPS Code of Ethics.

\item {\bf Broader impacts}
    \item[] Question: Does the paper discuss both potential positive societal impacts and negative societal impacts of the work performed?
    \item[] Answer: \answerYes{} 
    \item[] Justification: The paper includes a Broader Impacts section in Section \ref{app:Broader_Impacts}, which discusses potential impacts.
    
\item {\bf Safeguards}
    \item[] Question: Does the paper describe safeguards that have been put in place for responsible release of data or models that have a high risk for misuse (e.g., pre-trained language models, image generators, or scraped datasets)?
    \item[] Answer: \answerNA{} 
    \item[] Justification: The paper poses no such risks.

\item {\bf Licenses for existing assets}
    \item[] Question: Are the creators or original owners of assets (e.g., code, data, models), used in the paper, properly credited and are the license and terms of use explicitly mentioned and properly respected?
    \item[] Answer: \answerYes{} 
    \item[] Justification: The creators or original owners of assets (code, data, models), used in the paper, properly credited and are the license and terms of use explicitly mentioned and properly respected in Section \ref{sec:Experiments_setup} and Appendix \ref{app:supp_results}.

\item {\bf New assets}
    \item[] Question: Are new assets introduced in the paper well documented and is the documentation provided alongside the assets?
    \item[] Answer: \answerNA{} 
    \item[] Justification: The paper does not release new assets.

\item {\bf Crowdsourcing and research with human subjects}
    \item[] Question: For crowdsourcing experiments and research with human subjects, does the paper include the full text of instructions given to participants and screenshots, if applicable, as well as details about compensation (if any)? 
    \item[] Answer: \answerNA{} 
    \item[] Justification: The evaluation did not use crowdsourcing or external human participants. The model-output preference evaluation was conducted by the authors themselves.

\item {\bf Institutional review board (IRB) approvals or equivalent for research with human subjects}
    \item[] Question: Does the paper describe potential risks incurred by study participants, whether such risks were disclosed to the subjects, and whether Institutional Review Board (IRB) approvals (or an equivalent approval/review based on the requirements of your country or institution) were obtained?
    \item[] Answer: \answerNA{} 
    \item[] Justification: The evaluation did not involve external human participants; the model-output preference judgments were performed by the authors themselves. No personal, identifiable, or sensitive information was collected.

\item {\bf Declaration of LLM usage}
    \item[] Question: Does the paper describe the usage of LLMs if it is an important, original, or non-standard component of the core methods in this research? Note that if the LLM is used only for writing, editing, or formatting purposes and does \emph{not} impact the core methodology, scientific rigor, or originality of the research, declaration is not required.
    \item[] Answer: \answerNA{} 
    \item[] Justification: The LLM is used only for editing, or formatting purposes and does \emph{not} impact the core methodology, scientific rigor, or originality of the research. Thus, declaration is not required.

\end{enumerate}

\end{document}